%% file: paper.tex
\documentclass[letterpaper,twocolumn,10pt]{article}
\usepackage{usenix2019,epsfig,endnotes}
\usepackage{algorithm2e}
\usepackage{enumitem}
\usepackage{amsmath}
\usepackage{url}
\usepackage[title]{appendix}

\usepackage{tikz}
\newcommand*\circled[1]{\tikz[baseline=(char.base)]{
            \node[shape=circle,draw,inner sep=1pt] (char) {\textcolor[rgb]{.55,0,0} {#1}};}}

\newcommand{\localsgd}{\circled{1}}
\newcommand{\getnoise}{\circled{2}}
\newcommand{\noisyupdate}{\circled{3}}
\newcommand{\krum}{\circled{4}}
\newcommand{\signedupdate}{\circled{5}}
\newcommand{\shamir}{\circled{6}}
\newcommand{\secagg}{\circled{7}}
\newcommand{\blockgen}{\circled{8}}

\begin{document}

\date{}

\title{\Large \bf Biscotti: A Ledger for Private and Secure Peer-to-Peer Machine Learning}

\author{
{\rm Muhammad Shayan} \\
University of British Columbia \\
mdshayan@alumni.ubc.ca
\and
{\rm Clement Fung} \\
University of British Columbia \\
clement.fung@alumni.ubc.ca
\and
{\rm Chris J.M. Yoon} \\
University of British Columbia \\
yoon@alumni.ubc.ca
\and
{\rm Ivan Beschastnikh} \\
University of British Columbia \\
bestchai@cs.ubc.ca
}

\maketitle


\subsection*{Abstract}

Federated Learning is the current state of the art in supporting
secure multi-party machine learning (ML): data is maintained on the owner's device and the updates to the model are
aggregated through a secure protocol~\cite{Bonawitz:2017}. However, this process assumes a
trusted centralized infrastructure for coordination, and clients must
trust that the central service does not use the byproducts of client data. In addition to this, a group of malicious clients could also harm the performance of the model by carrying out a poisoning attack.~\cite{Huang:2011}

As a response, we propose Biscotti: a fully decentralized peer to peer (P2P) approach
to multi-party ML, which uses blockchain and cryptographic primitives to coordinate
a privacy-preserving ML process between peering clients. Our
evaluation demonstrates that Biscotti is scalable, fault tolerant, and
defends against known attacks. For example, Biscotti is able to protect the privacy of an individual client's update and the performance of the global model at scale when 30\% of
adversaries are trying to poison the model~\cite{Huang:2011}.

\input{intro}
\input{challenges}
\input{threatmodel}
\input{design}

\input{implementation}

\input{eval}
\input{discussion}
\input{related}

\input{conc}

\subsection*{Acknowledgments}

We would like to express our thanks to Margo Seltzer for her helpful feedback and comments. We would like to further thank Matheus Stolet for his help in evaluating the prototype at scale, and Heming Zhang for his help with bootstrapping the deployment of Biscotti on PyTorch. We would also like to thank Gleb Naumenko, Nico Ritschel, Jodi Spacek, and Michael Hou for their work on the initial Biscotti prototype. 

This research has been sponsored by the Huawei Innovation Research
Program (HIRP), Project No: HO2018085305. We also acknowledge the
support of the Natural Sciences and Engineering Research Council of
Canada (NSERC), 2014-04870.


{\footnotesize \bibliographystyle{acm}
\bibliography{paper}}

\begin{appendices}
  \input{appendix}
\end{appendices}

\end{document}

%% file: intro.tex
\section{Introduction}
\label{sec:intro}

A common thread in machine learning (ML) applications is the collection of massive
amounts of training data, which is often centralized for analysis.
However, when training ML models in a multi-party setting, users must
share their potentially sensitive information with a centralized
service. 



Federated learning~\cite{Bonawitz:2017, McMahan:2017} is a prominent solution for high scale secure multi-party ML: clients train a shared
model with a secure aggregator without revealing their underlying data
or computation.  But, doing so introduces a subtle threat: clients,
who previously acted as passive data contributors, are now actively
involved in the training process~\cite{Huang:2011}. This presents new
privacy and security challenges.


Prior work has demonstrated that adversaries can attack the shared
model through poisoning attacks~\cite{Biggio:2012, Nelson:2008}, in
which an adversary contributes adversarial updates to influence shared
model parameters. Adversaries can also attack the privacy of
other clients in federated learning. In an information leakage attack,
an adversary poses as an honest client and attempts to steal or
de-anonymize sensitive training data through careful observation and
isolation of a victim's model updates~\cite{Hitaj:2017, Melis:2018}.

Solutions to these two attacks are
at tension and are inherently difficult to achieve together: client
contributions or data can be made public and verifiable to prevent
poisoning, but this violates the privacy guarantees of federated
learning. Client contributions can be made more private, but this
eliminates the potential for accountability from adversaries. Prior
work has attempted to solve these two attacks individually through
centralized anomaly detection~\cite{Shen:2016}, differential
privacy~\cite{Abadi:2016, Dwork:2014, Geyer:2017} or secure
aggregation~\cite{Bonawitz:2017}. 
However, a private and decentralized solution that solves both attacks
concurrently does not yet exist. In addition, these approaches are
inapplicable in contexts where a trusted centralized authority does
not exist. This is the focus of our work.

Because ML does not require strong consensus or consistency to
converge~\cite{Recht:2011}, traditional strong consensus protocols
such as Byzantine Fault Tolerant (BFT) protocols~\cite{Castro:1999} are too restrictive for machine learning
workloads. To facilitate private, verifiable, crowd-sourced
computation, distributed ledgers (blockchains)~\cite{Nakamoto:2009}
have emerged.
Through design elements, such as publicly verifiable proof of work,
eventual consistency, and ledger-based consensus, blockchains have
been used for a variety of decentralized multi-party tasks such as
currency management~\cite{Gilad:2017, Nakamoto:2009}, archival data
storage~\cite{Azaria:2016, Miller:2014} and financial
auditing~\cite{Narula:2018}. Despite this wide range of applications,
a fully realized, accessible system for large scale multi-party ML
that is robust to both attacks on the global model and attacks on
other clients does not exist. 



We propose \emph{Biscotti}, a decentralized public peer to peer system that co-designs
a privacy-preserving multi-party ML process with a blockchain ledger.
Peering clients join Biscotti and contribute to a ledger to train a
global model, under the assumption that peers are willing to
collaborate on building ML models, but are unwilling to share their
data. Each peer has a local data set that could be used to train the model.
Biscotti is designed to support stochastic gradient descent
(SGD)~\cite{Bottou:2010}, an optimization algorithm that iteratively
selects a batch of training examples, computes their gradients with respect to
the current model parameters, and takes steps in the
direction that minimizes the loss function. SGD is general purpose and
can be used to train a variety of models, including neural networks~\cite{Dean:2012}.

The Biscotti blockchain coordinates ML training between the
peers.  
Peers in the system are weighed by the value, or stake, that they have
in the system. Inspired by prior work~\cite{Gilad:2017}, Biscotti uses consistent hashing based on
proof of stake in combination with verifiable random functions
(VRFs)~\cite{Micali:1999} to select key roles that help to arrange the
privacy and security of peer SGD updates.
Our use of stake prevents groups of colluding peers from overtaking
the system without a sufficient stake ownership. 


\emph{With Biscotti's design our primary contribution is to combine several
prior techniques into one coherent system that provides secure and
private multi-party machine learning in a highly distributed
setting.} In particular, Biscotti prevents peers from poisoning the
model through the Multi-KRUM defense~\cite{Blanchard:2017} and
provides privacy through differentially private
noise~\cite{Abadi:2016} and Shamir secrets for secure
aggregation~\cite{Shamir:1979}.
Most importantly, Biscotti combines these techniques \emph{without}
impacting the accuracy of the final model, achieving identical model
performance as federated learning.









We evaluated Biscotti on Azure and considered its
performance, scalability, churn tolerance, and ability to withstand
different attacks.
We found that Biscotti can train an MNIST softmax model with 200 peers
on a 60,000 image dataset in 266.7 minutes, which is {14x}
slower than a similar federated learning deployment. Biscotti is fault tolerant
to node churn every 15 seconds across 100 nodes, and converges even
with such churn.
Furthermore, we show that Biscotti is resilient to information leakage
attacks~\cite{Melis:2018} that require knowledge of a client's SGD
update and a label-flipping poisoning attack~\cite{Huang:2011} from
prior work.


%% file: challenges.tex
\section{Challenges and contributions}
\label{sec:challenges}

  We now describe the key challenges in designing a peer
  to peer (P2P) solution for multi-party ML and the key pieces in
  Biscotti's design that resolve each of these challenges. \vspace{.1em}

  \noindent \textbf{Sybil attacks: Consistent hashing using VRF's and proof of stake.}

  In a P2P setting adversaries can collude or generate aliases to increase their
  influence in a sybil attack~\cite{Douceur:2002}.



  Biscotti uses a consistent hashing protocol based on the hash of the last block and verifiable random functions (VRF's) (see Appendix ~\ref{sec:vrf}) to select a subset of peers that are responsible 
  for the different stages of the training process: adding noise to
  updates, validating an update, and securely aggregating the update. To mitigate the effect of sybils, the protocol selects peers proportionally to peer stake.  A peer's stake is reputation that the peer acquires by contributing in the system. This ensures that an adversary cannot
  increase their influence in the system by creating multiple peers without increasing their total contribution/
  stake. \vspace{.1em}


   \noindent \textbf{Poisoning attacks: update validation using KRUM.}
  In multi-party ML, peers possess a relatively disjoint and private sub-set of the
  training data. As mentioned above, privacy can be exploited by adversaries to provide cover for poisoning attacks.

  In multi-party settings, known baseline models are not available to
  peers, so Biscotti validates an SGD update by evaluating an update
  with respect to the updates submitted by other peers. Biscotti
  validates an SGD update using the \emph{Multi-KRUM}
  algorithm~\cite{Blanchard:2017}, which rejects updates that push the
  model away from the direction of the majority of the updates. More
  precisely, in each round, a committee of peers is selected to filter
  out poisoned updates by a majority vote and each member of the
  committee uses Multi-KRUM to filter out poisoned SGD updates. Multi-KRUM guarantees to filter out \emph{f} out of \emph{n} malicious updates such that $2f + 2 < n$. We demonstrate that by using Multi-KRUM, Biscotti can handle poisonous updates from up to 30\% malicious clients.

  \noindent \textbf{Information leakage attacks: random verifier peers
    and differentially private updates using pre-committed noise.}
  %
  By observing a peer's SGD updates from each verification round,
  an adversary can perform an information leakage
  attack~\cite{Melis:2018} and recover details about a victim's training
  data. (For details on information leakage attacks, see Appendix ~\ref{sec:infoleakage})

  Biscotti prevents such attacks during update verification in two ways. First,
  it uses a consistent hashing on the SHA-256 hash of the last block to
  ensure that malicious peers cannot deterministically select themselves to 
  verify a victim's gradient. Second, peers send differentially-private
  updates~\cite{Abadi:2016, Dwork:2014} to verifier peers:
  before sending a gradient to a
  verifier, pre-committed $\varepsilon$-differentially private noise is
  added to the update, masking the peer's gradient in a way that neither
  the peer nor the attacker can influence or observe(See Figure \ref{fig:evalprivacy}). By verifying
  noised SGD updates, peers in Biscotti can verify the updates of other
  peers without observing their un-noised versions. \vspace{.1em} 

  \noindent \textbf{Utility loss with differential privacy: secure
    update aggregation and cryptographic commitments.}
  The blockchain-based ledger of model updates allows for auditing of
  state, but this transparency is counter to the goal of
  privacy-preserving multi-party ML. For example, the ledger trivially
  leaks information if we store SGD updates directly.

  Verification of differentially private updates discussed above is one
  piece of the puzzle. The second piece is secure update aggregation: a
  block in Biscotti does not store updates from individual peers, but
  rather an aggregate that obfuscates any single peer's contribution
  during that round of SGD. Biscotti uses a verifiable secret sharing
  scheme~\cite{Aniket:2010} to aggregate the updates so that any
  individual update remains hidden through cryptographic commitments (For a background, see Appendix ~\ref{sec:crypto}).

  \begin{figure}[t]
    \centering
    \includegraphics[width=\linewidth]{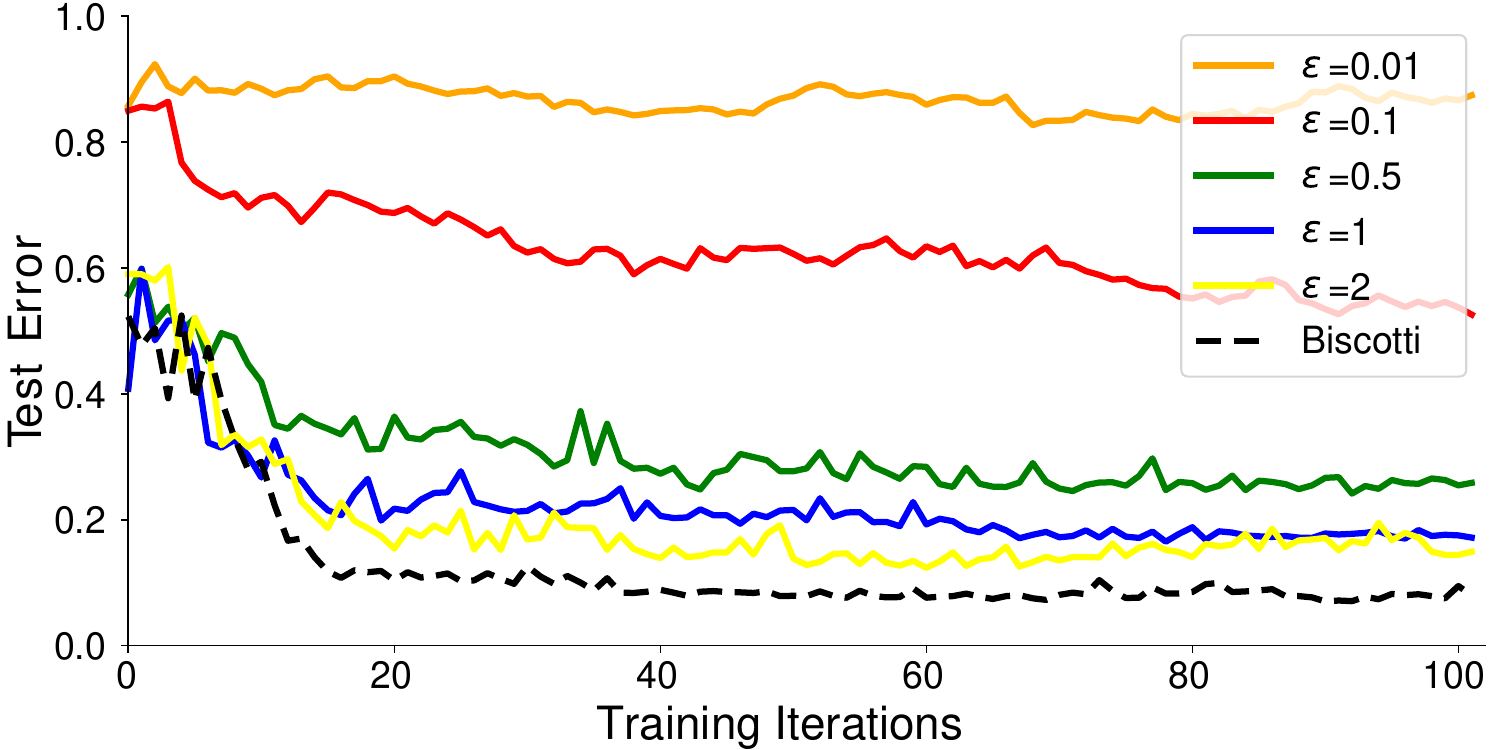}
    \caption{The utility trade-off if differentially private noise is
      added directly to the updates in the ledger. Biscotti, which
      aggregates non-noisy updates, has the best utility (lowest
      test error).
    }
    \label{fig:privacyutility}
  \end{figure}

  \begin{figure}[t]
    \centering
    \includegraphics[width=\linewidth]{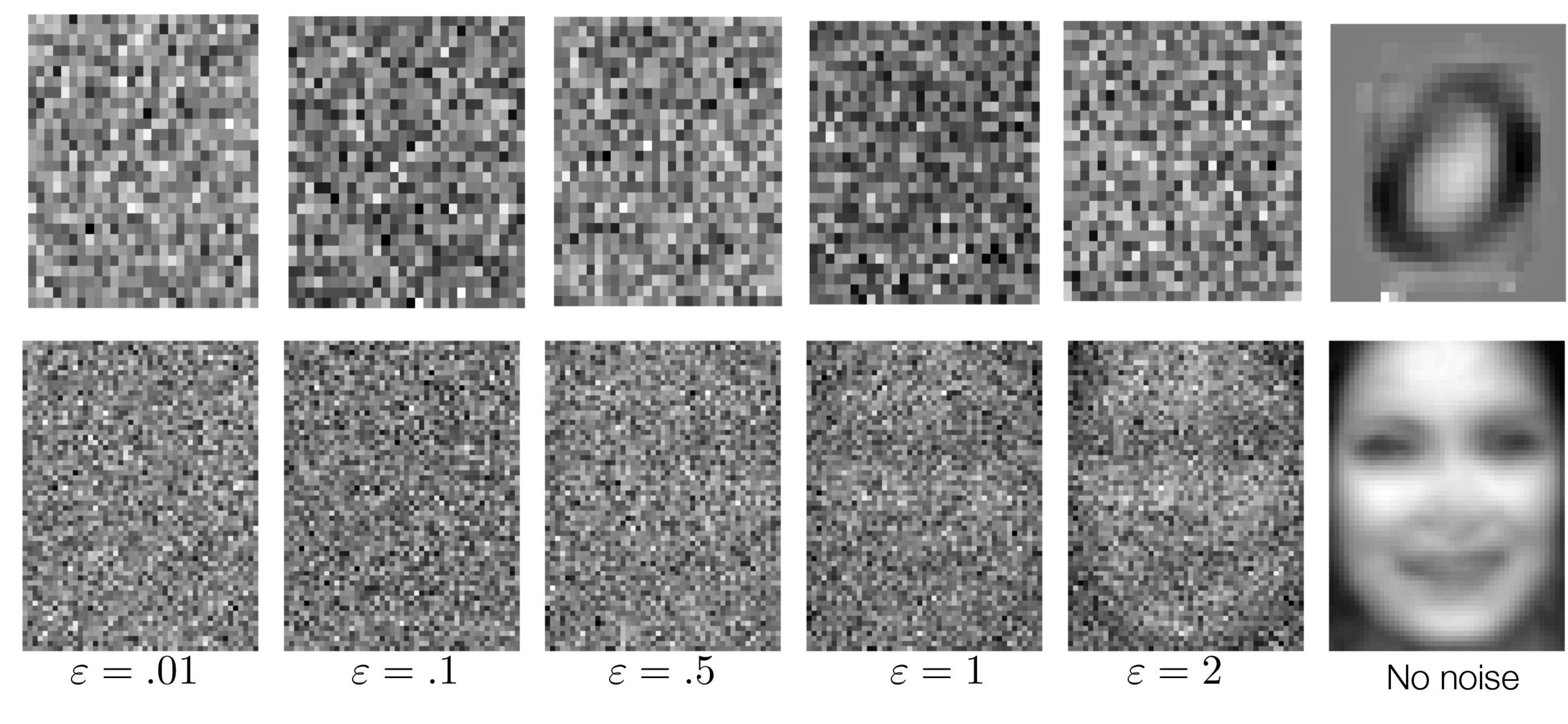}
    \caption{Visualized privacy trade-off for curves in
      Figure~\ref{fig:privacyutility}.}
    \label{fig:evalprivacy}
  \end{figure}

  However, secure aggregation can be either done on the
  differentially-private updates or the original updates with no noise. 
  This design choice represents a privacy-utility tradeoff in the trained model.
  By aggregating differentially-private updates the noise is pushed into
  the ledger and the final model has lower utility. By aggregating
  original updates we can achieve utility that is identical to federated
  learning training. Figure~\ref{fig:privacyutility} illustrates this
  trade-off for learning a softmax model to recognize hand-written digits. The model is trained using the MNIST dataset over 100 iterations for different values of $\epsilon$. Smaller
  $\epsilon$ means more noise and more privacy, but lower utility, or
  test error. The bottom-most line in the Figure is Biscotti,
  which aggregates original updates: a design choice that we have made
  to prioritize utility. 
  By default Biscotti uses a batch value of
  35. Figure~\ref{fig:evalprivacy} illustrates the privacy-utility
  trade-off visually for batches of size 35 with noisy updates as compared to
  Biscotti (right-most image), which aggregates un-noised updates.
  The
  images are reconstructed using an information leakage attack~\cite{Melis:2018} on the
  aggregated gradients of two different machine learning models. The top row of images are constructed from aggregated gradients of the MNIST model with respect to the 0 digit class. The bottom row shows results for a softmax model trained to recognize gender from faces using the Labelled Faces in the Wild (LFW) dataset. The pictures are reconstructed with respect to the female class. These results demonstrate that although aggregation protects individual training examples it does reveal information about how a certain class looks like on average in the aggregate. Differential private noise needs to be included in the aggregate to provide this additional level of protection. However, our design favours utility, therefore we choose to remove the noise before aggregating the updates in the ledger.

%% file: threatmodel.tex
\section{Assumptions and threat model}
\label{sec:threatmodel}




Like federated learning, Biscotti assumes a public machine learning system which peers can join/leave anytime. Biscotti assumes that users are willing to
collaborate on building ML models, but are unwilling
to directly share their data when doing so~\cite{McMahan:2017}. This need for collaboration arises because a model trained only on an individual peer's data would exhibit poor performance, but a model trained by
all participants will have near-optimal performance.

\subsection{Design assumptions}

\noindent \textbf{Proof of stake}. 

Peers in Biscotti need to arrive at a consensus on the state of the model for each round of training. Proof of Stake (POS)~\cite{Gilad:2017,Aggelos:2017}  is a recently proposed consensus mechanism for cryptocurrencies. Consensus using POS is fast because it delegates the consensus responsibility to a subset of peers each round. The peers responsible are selected each round based on the amount of stake that they possess in the system. In cryptocurrencies, the \emph{stake} of a peer refers to the amount of value (money) that a peer holds. POS relies on the assumption that a subset of peers holding a significant fraction of the stake will not attempt to subvert the system. For a detailed background on Proof of Stake , please refer to Appendix ~\ref{sec:stake}.  

In Biscotti, we define \emph{stake} as a measure of peer's contribution to the system. Peers acquire stake by by providing SGD updates or by facilitating the consensus process. Thus, the stake that a peer accrues during the training process is proportional to their contribution to the trained model. We assume that all nodes can obtain the fraction of stake of any other peer using a function from the current stake of the blockchain. In addition to this, we also assume that at any point in time more than 70\% of the stake in the system is honest and that the stake function is properly bootstrapped. The initial stake distribution at the start may be derived from an online data sharing marketplace, a shared reputation score among competing agencies or auxiliary information on a social network.

\noindent \textbf{Blockchain topology}. Each peer is connected to some
subset of other peers in a topology that allows flooding-based
dissemination of blocks that eventually reach all peers. For example,
this could be a random mesh topology with flooding, similar to the one
used for block dissemination in
Bitcoin~\cite{Nakamoto:2009}. Peers that go offline during training and join later are bootstrapped on the current state of the blockchain by other peers in the system. \vspace{.1em}


\noindent \textbf{Machine learning}. We assume that ML training
parameters are known to all peers: the model, its hyperparameters, its
optimization algorithm and the learning goal of the system (these are
distributed in the first block). In a non-adversarial setting, peers
have local datasets that they wish to keep private. When peers draw a
sample from this data to compute an SGD update, we assume that this is
done uniformly and independently~\cite{Bottou:1999}. This ensures that
the Multi-KRUM gradient validation~\cite{Blanchard:2017} is accurate.


\subsection{Attacker assumptions}

Peers may be adversarial and send malicious updates to perform a
poisoning attack on the shared model or an information leakage attack
against a target peer victim. In doing so, we assume that the
adversary may control multiple peers in a sybil
attack~\cite{Douceur:2002} but does not control more than 30\% of the total peers. Although peers may
be able to increase the number of peers they control in the system, we
assume that adversaries cannot artificially increase their stake in
the system except by providing valid updates that pass
Multi-KRUM~\cite{Blanchard:2017}. 


When adversaries perform a \emph{poisoning attack}, we assume that
their goal is to harm the performance of the final global model. Our
defense relies on filtering out malicious updates that are
sufficiently different from the honest clients and are pushing the
global model towards some sub-optimal objective. For the purposes of this work, we limit the adversaries only to a label flipping poisoning attack ~\cite{Huang:2011} in which they mislabel a certain class in their dataset causing the model to learn to misclassify it. 
This does not include attacks on unused parts of the model topology,
like backdoor attacks~\cite{Bagdasaryan:2018}, attacks based on gradient-ascent~\cite{Munoz-Gonzalez:2017} techniques or adaptive attacks based on knowledge of the poisoning defense ~\cite{Bhagoji:2019}.

When adversaries perform an \emph{information leakage attack}, we
assume that they aim to learn properties of a victim's local dataset.
Due to the vulnerabilities of secure aggregation, we do not consider information leakage attacks with side information~\cite{Fredrikson:2015,Salem:2018} and class-level information leakage
attacks on federated learning~\cite{Hitaj:2017}, which attempt to
learn the properties of an entire target class. 

%% file: design.tex
\section{Biscotti design} \label{sec:design}

\begin{figure*}[t]
  \centering
  \includegraphics[width=1.0\linewidth]{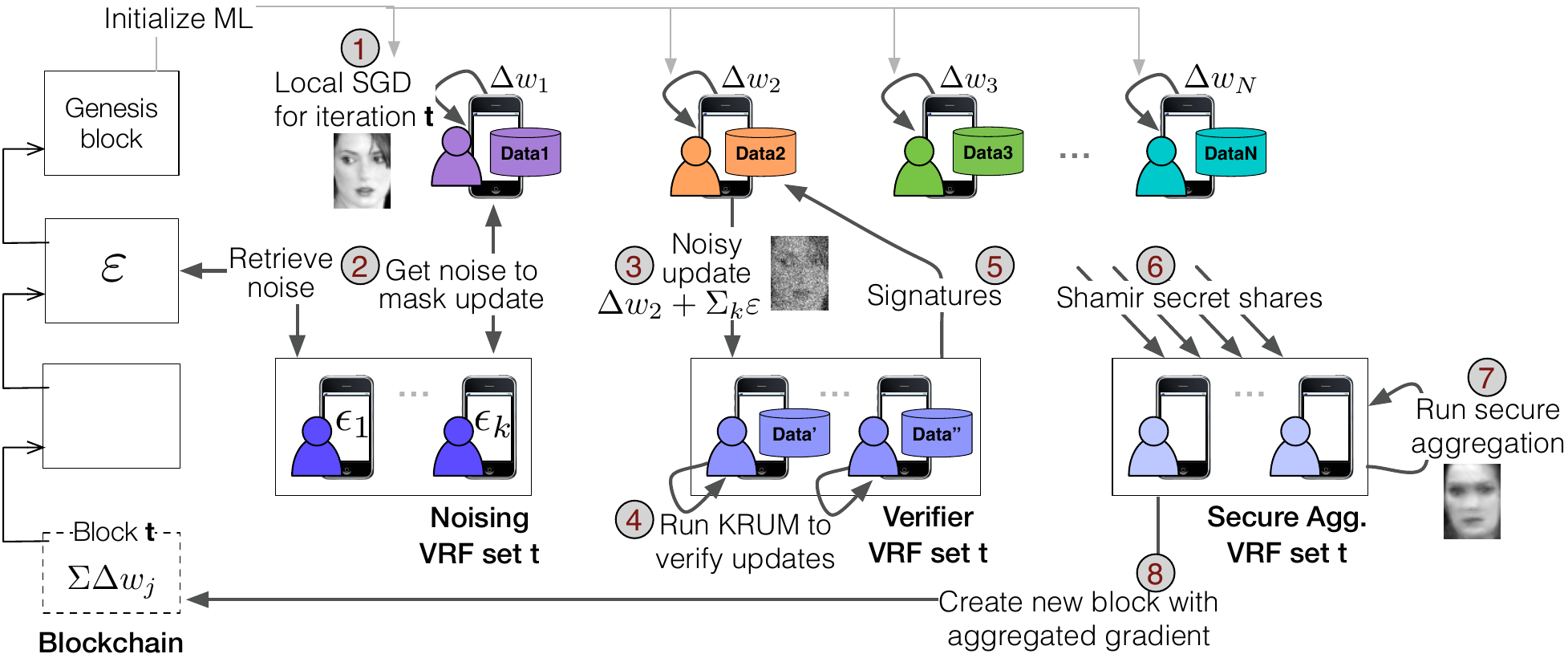}
  \caption{The ordered steps in a single iteration of the Biscotti
  algorithm.
    \label{fig:design}}
\end{figure*}

Biscotti implements peer-to-peer ML trained using SGD. For this
process, Biscotti's design has the following goals:

\begin{itemize}[nosep, wide]

  \item Convergence to an optimal global model (the model trained without adversaries in a federated learning setting) 

  \item Poisoning is prevented by verifying peer contributions to the model

  \item Peer training data is kept private and information leakage
    attacks on training data are prevented

  \item Colluding peers cannot gain influence without acquiring
    sufficient stake

\end{itemize}

Biscotti meets these goals through a blockchain-based design
that we describe in this section. \vspace{.1em}

\noindent \textbf{Design overview.}
Peers join Biscotti and
collaboratively train a global model. Each block in the distributed
ledger represents a single iteration of SGD and the ledger contains
the state of the global model at each iteration. 
Figure~\ref{fig:design} overviews the Biscotti design with a
step-by-step of illustration of what happens during a \emph{single}
SGD iteration in which a single block is generated. 

In each iteration, peers locally compute SGD updates (step 
\localsgd{} in Figure~\ref{fig:design}). Since SGD updates need to be
kept private, each peer first \emph{masks} their update using
differentially private noise. This noise is obtained from a unique set
of noising peers for each client selected by a VRF~\cite{Micali:1999} (step \getnoise{} and \noisyupdate). 


The masked updates are validated by a verification committee to defend against poisoning. Each member in the verification committee signs the commitment to the peer's \emph{unmasked} update if it passes Multi-KRUM (step \krum). If the majority of the committee signs an update (step \signedupdate), the signed update is divided into Shamir secret shares (step \shamir) and given to a aggregation committee. The aggregation committee uses a secure protocol to aggregate the unmasked updates (step \secagg). All peers who contribute a share to the final update along with the peers chosen for the verification and aggregation committees receive additional stake in the system. 


%
The aggregate of the updates is added to the global model in a newly created block which is disseminated to all the peers and appended to the ledger (step \blockgen). Using the updated global model and stake, the peers repeat (step \localsgd). 

Next, we describe how we bootstrap the training process.



\subsection{Initializing the training}

Biscotti peers initialize the training process using information in
the first (genesis) block. We assume that this block is distributed
out of band by a trusted authority and the information within it is
reliable. The centralized trusted authority only plays the role of facilitating and bootstrapping the training process by distributing the genesis block that makes public information available to all peers in the system. It is not entrusted with individual SGD updates of the peers that could potentially leak private information of a peer's data. Each peer that joins Biscotti obtains the following
information from the genesis block:

\begin{itemize}[nosep, wide]
  \item Initial model state $w_0$, and expected number of iterations $T$
  \item The commitment public key $PK$ for creating commitments to SGD
  updates (see Appendix~\ref{sec:crypto})
  \item The public keys $PK_{i}$ of all other peers in the system that
  are used to create and verify signatures during verification. 
  \item Pre-commitments to $T$ iterations of differentially private
  noise $\zeta_{1..T}$ for each SGD iteration by each peer (see
  Figure~\ref{fig:vrfnoise} and Appendix~\ref{sec:diffpriv})
  \item The initial stake distribution among the peers
  \item A stake update function for updating a peer's stake when a
  new block is appended
\end{itemize}





\subsection{Blockchain design}
\label{subsec:blockchaindesign}

\begin{figure}[t]
  \centering
  \includegraphics[width=\linewidth]{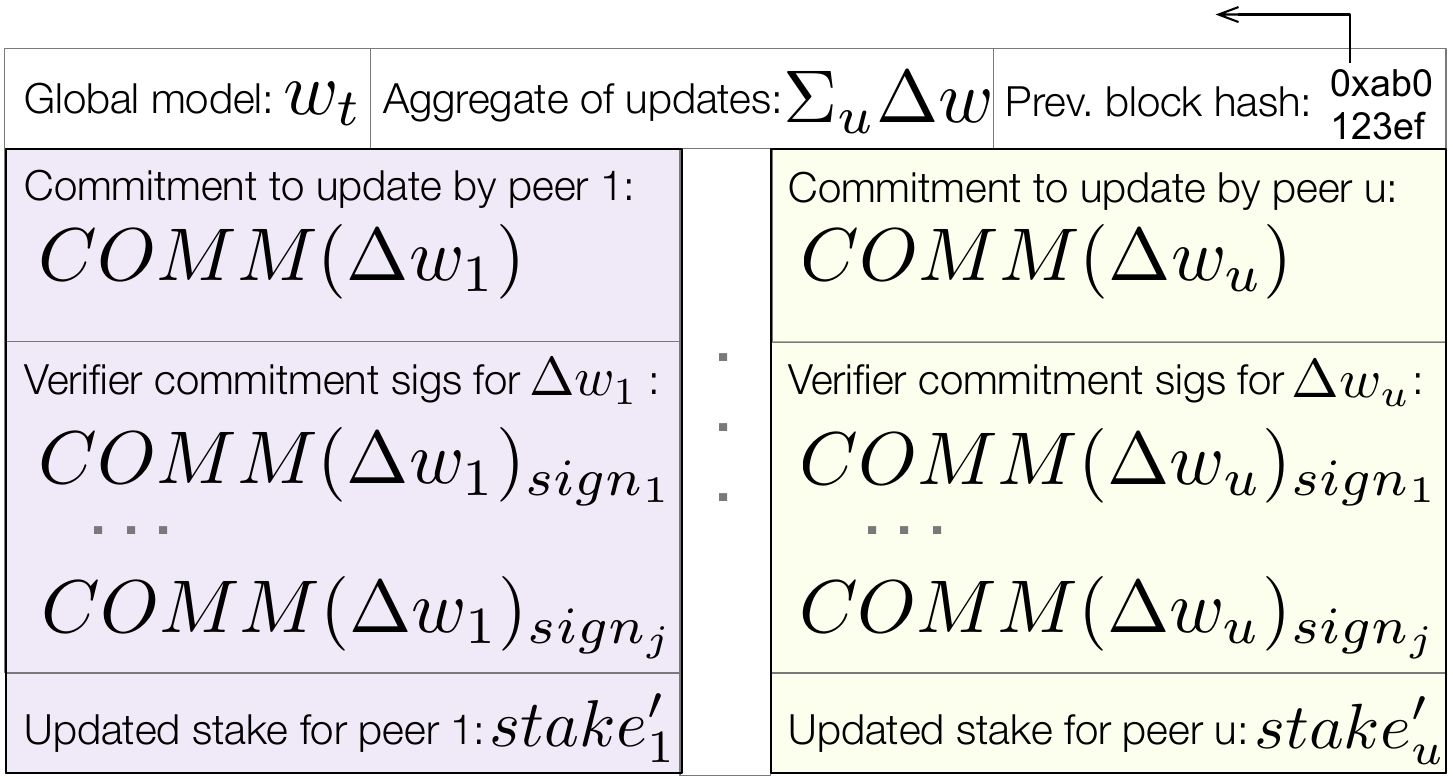}
  \caption{Block contents at iteration $t$. Note that $w_t$ is
    computed using $w_{t-1} + \Sigma w_u$ where $w_{t-1}$ is the
    global model stored in the block at iteration $t-1$ and $j$ is
    the set of verifiers for iteration $t$.
    \label{fig:block}}
\end{figure}

Distributed ledgers are constructed by appending read-only blocks to a
chain structure and disseminating blocks using a gossip protocol.
Each block maintains a pointer to its previous block as a
cryptographic hash of the contents in that block. 






Each block in Biscotti (Figure ~\ref{fig:block}) contains, in addition to the previous block
hash pointer, an aggregate ($\Delta w$) of SGD updates from multiple
peers and a snapshot of the global model $w_t$ at iteration $t$. Newly
appended blocks to the ledger store the aggregated updates $\sum
\Delta w_i$ of multiple peers. To verify that the aggregate was honestly computed, individual updates need to be made part of the block. However, storing them would leak information about private training data of the individuals. We solve this problem by using polynomial commitments.~\cite{Aniket:2010}. Polynomial commitments take an SGD update and map it to a point on an elliptic curve. (see Appendix ~\ref{sec:crypto} for details). By including a list of 
commitments for each peer $i$'s update $COMM~(\Delta w_i)$ in the block, we can provide both privacy and verifiability of the aggregate. The commitments provide privacy by hiding the
individual updates yet can be homomorphically combined to verify that
the update to the global model by the aggregator $\sum \Delta w_i$ was computed honestly. The following equality holds if the list of
committed updates equals the aggregate sum:
  \begin{align*}
    COMM( \sum \Delta w_i) = \prod_{i} {COMM( \Delta w_i)} &&&& (2)
  \end{align*}

%
The training process continues for a specified number of iterations $T$ upon which the learning process is terminated and each peer
extracts the global model from the final block.


%

\subsection{Using stake for role selection}
\label{sec:peerselection}

\begin{figure}[t]   
\centering 
\includegraphics[width=.85\linewidth]{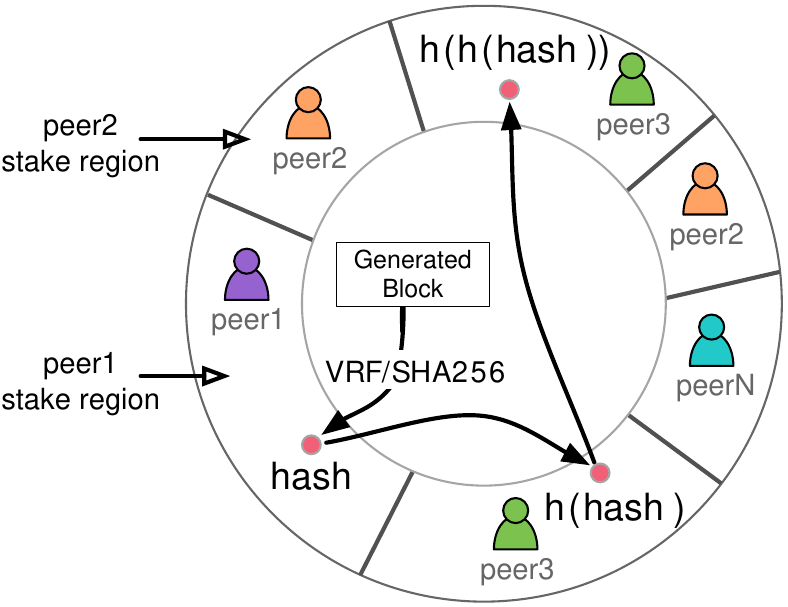} 
\caption{Biscotti uses a consistent hashing protocol based on the
  current stake to determine the roles for each iteration.
  \label{fig:stakering} 
}
\end{figure}

For each iteration in Biscotti, a consistent hashing protocol  weighted by stake designates roles (noiser, verifier, aggregator)
to some peers in the system. The protocol ensures that the influence of a peer is bounded by their stake (i.e. adversaries cannot trivially increase their influence through sybils). Peers can take on multiple roles in a given iteration but cannot be a verifier and aggregator in the same round. The verification and aggregation committees are the same for all peers but the noising committee is unique to each peer. 

Biscotti uses consistent hashing to select peers for the verification and aggregation committees
(Figure~\ref{fig:stakering}). The initial SHA-256 hash of the last block is repeatedly
re-hashed: each new hash result is mapped onto a hash ring where
portions of the ring are proportionally assigned to peers based on their stake. The peer in whose space the hash lies is chosen for the verifier/aggregator role. The process is repeated until you get verifier/aggregator committees of the right size. This provides the same stake-based property as Algorand~\cite{Gilad:2017}: a peer's probability of winning this lottery is proportional to their stake. Since an adversary cannot predict the future state of a block until it is created, they cannot speculate on outputs of consistent hashing and strategically perform attacks. 

Unlike verification and aggregation, the noising committee is  different for each peer for additional privacy. Each peer can arrive at a unique committee for itself via consistent hashing by using a different initial hash. The hash computed should be random yet globally verifiable by other peers in the system. In Biscotti, a peer computes this hash by passing their secret key $SK_{i}$ and the SHA-256 hash of the last block to a verifiable random function (VRF). By virtue of the peer's secret key, the hash computed is unique to the peer resulting in distinct committees for different peers. The VRF hash is also accompanied by a proof that can be combined with a peer’s public key allowing others
peer to determine that the correct noise providers are selected by the peer. 

At each iteration, peers run the consistent hashing protocol to determine whether they are an aggregator or verifier. Peers that are part of the verification and aggregation committees do not contribute updates for that round. Each peer that is not an aggregator or verifier computes their set of noise providers. The contributing peers then obtain noise to hide their updates by following the noising protocol.

\subsection{Noising protocol} \label{sec:noise}

Sending a peer's SGD update directly to another peer may leak sensitive information about their private dataset~\cite{Melis:2018}. To
prevent this, peers use differential privacy to hide their updates
prior to verification by adding noise sampled from a normal distribution. This ensures that each step is ($\epsilon,\delta$) differentially private by standard arguments in ~\cite{Dwork:2014}. 
(see Appendix~\ref{sec:diffpriv} for formalisms).



\noindent \textbf{Using pre-committed noise to thwart poisoning.}
Attackers may maliciously use the noising protocol to execute
poisoning or information leakage attacks. For example, a peer can send
a poisonous update $\Delta w_{poison}$, and add noise $\zeta_p$
that \emph{unpoisons} this update to resemble an honest update $\Delta w$,
such that $\Delta w_{poison} + \zeta_{p} = \Delta w$. By doing
this, a verifier observes the honest update $\Delta w$, but the
poisonous update $\Delta w_{poison}$ is applied to the model because the noise is removed in the final aggregate.

To prevent this, Biscotti requires that every peer pre-commits the
noise vector $\zeta_t$ for every iteration $t \in [1 .. T]$ in the
genesis block. Since the updates cannot be generated in advance
without the knowledge of the global model, a peer cannot
effectively commit noise that unpoisons an update. Furthermore,
Biscotti requires that the noise that is added is taken from a
\emph{different} peer than the one creating the update. This peer is
determined using a noising VRF and further restricts the control that
a malicious peer has over the noise used to sneak poisoned updates past
verification. \\

\noindent \textbf{Using a VRF-chosen set of noisers to thwart
information leakage.} A further issue may occur in which a noising
peer $A$ and a verifier $B$ can collude in an information leakage
attack against a victim peer $C$. The noising peer $A$ can commit a
set of zero noise that does not hide the original gradient value at
all. When the victim peer $C$ sends its noised gradient to the
verifier $B$, $B$ performs an information leakage attack on client $C$'s gradient back to its original training data. This
attack is viable because the verifier $B$ knows that $A$ is providing
the random noise and $A$ provides zero noise.

This attack motivates Biscotti's use of a private VRF that selects a
\emph{group} of noising peers based on the victim $C$'s secret key. In
doing so, an adversary cannot pre-determine whether their noise will
be used in a specific verification round by a particular victim, and
also cannot pre-determine if the other peers in the noising committee
will be malicious in a particular round. Our results in 
Figure~\ref{fig:evalnoiseattack} show that the probability of an
information leakage is negligible given a sufficient number of
noising peers. \\

\begin{figure}[t]   
\centering
\includegraphics[width=1.0\linewidth]{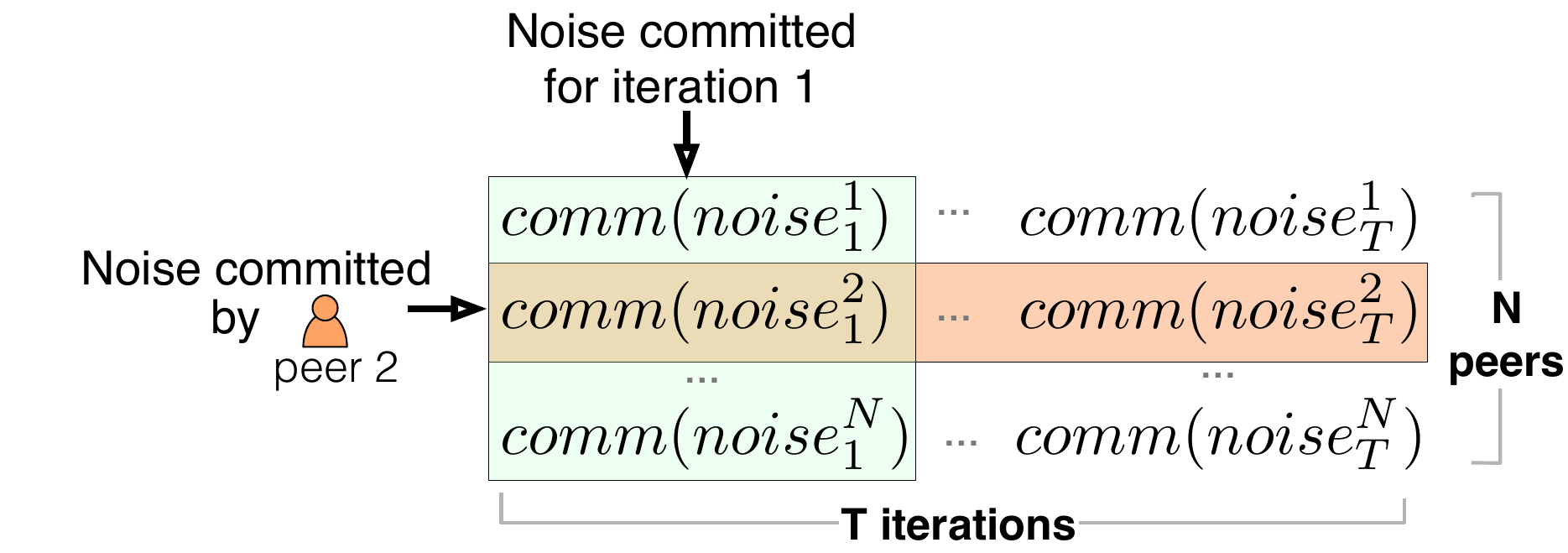}
\caption{Peers commit noise to an $N$ by $T$ structure. Each row $i$
  contains all the noise committed by a single peer $i$, and each
  column $t$ contains potential noise to be used during iteration $t$.
  When committing noise at an iteration $i$, peers execute a VRF and
  request $\zeta_i^k$ from peer $k$}
\label{fig:vrfnoise} 
\end{figure}

\noindent \textbf{Protocol description.} For an ML workload that may
be expected to run for a specific number of iterations $T$, each peer
$i$ generates $T$ noise vectors $\zeta_t$ and commits these noise
vectors into the ledger, storing a table of size $N$ by $T$ (Figure~
\ref{fig:vrfnoise}). When a peer is ready to contribute an update in
an iteration, it runs the noising VRF and contacts each noising peer
$k$, requesting the noise vector $\zeta_{k}$ pre-committed in the
genesis block $COMM(\zeta_{k}$). The peer then uses a verifier VRF to
determine the set of verifiers. The peer \emph{masks} their update
using this noise and submits to these verifiers the \emph{masked}
update, a commitment to the noise, and a commitment to the 
\emph{unmasked} update. It also submits the noise VRF proof that
attests to the verifier that its noise is sourced from peers that
are a part of their noise VRF set.

\subsection{Verification protocol}
\label{subsec:verification}

The verifier peers are responsible for filtering out malicious updates in a round by running Multi-KRUM on the received pool of updates and accepting the top majority of the updates received each round. Each verifier receives the following from each
peer $i$:
\begin{itemize} [nosep, wide]
  \item The masked SGD update: $(\Delta w_i + \sum_{k} \zeta_{k})$
  \item Commitment to the SGD update: $COMM(\Delta w_i)$ 
  \item The set of $k$ noise commitments:\\
  $\{COMM(\zeta_{1}), COMM(\zeta_{2}),..., COMM(\zeta_{k}))
  \}$ 
  \item A VRF proof confirming the identity of the $k$ noise peers
\end{itemize}

When a verifier receives a masked update from another peer, it can
confirm that the \emph{masked} SGD update is consistent with the
commitments to the \emph{unmasked} update and the noise by using the
homomorphic property of the commitments~\cite{Aniket:2010}. A
masked update is legitimate if the following equality holds:

  \begin{align*}
    COMM(\Delta w_i + \sum_{k} \zeta_{k}) = COMM(\Delta w_i)*\prod_{k} {COMM( \Delta \zeta_{k})}
  \end{align*}


Once the verifier receives a sufficiently large number of updates $R$,
it proceeds with selecting the best updates using Multi-KRUM, as follows:

\begin{itemize}

  \item For every peer $i$, the verifier calculates a score $s(i)$
    which is the sum of euclidean distances of $i$'s update to the
    closest $R-f-2$ updates. It is given by:
  
  \begin{center}
   $s(i)=\sum_{i\rightarrow j} \|(\Delta w_{i} + \sum_{i,k} \zeta_{i,k})- (\Delta w_{j}\ + \sum_{j,k} \zeta_{j,k})|^2$
   \end{center}

  where $i \rightarrow j$ denotes the fact that $(\Delta w_{j}\ + \sum_{j,k} \zeta_{j,k})$ belongs to the $R-f-2$ closest updates to $(\Delta w_{i} + \sum_{i,k} \zeta_{i,k})$.

 \item The $R-f$ peers with the lowest scores are selected while the rest are rejected. 

 \item The verifier signs the $COMM(\Delta w_i)$ using its public key for all the updates that are accepted.

\end{itemize} 

To prevent a malicious sybil verifier from accepting all updates of its colluders in this stage, we require a peer to obtain signatures from majority of the verifiers to get their update accepted. Once a peer receives a majority number of signatures from the verifiers, the update can be disseminated for aggregation.



\subsection{Aggregation protocol}

All peers with a sufficient number of signatures in the verification
stage submit their SGD updates for aggregation which will be eventually appended to the global model. The update equation in SGD (see 
Appendix~\ref{sec:background}) can be re-written as:
\begin{align*}
w_{t+1} = w_t + \sum_{i=1}^{\Delta w_{verified}}\Delta w_{i}
\end{align*}
where $\Delta w_{i}$ is the verified SGD update of peer $i$ and $w_
{t}$ is the global model at iteration $t$.

However, individual updates contain sensitive information and cannot be directly shared for aggregation. This presents a privacy
dilemma: no peer should observe an update from any other peer, but the sum of the
updates must be stored in a block.

The objective of the aggregation protocol is to enable a set of $m$
aggregators, predetermined by consistent hashing, to compute $\sum_
{i}\Delta w_{i}$ without observing any individual updates. Biscotti
uses a technique that preserves privacy of the individual updates
if at least half of the $m$ aggregators participate honestly in the
aggregation phase. This guarantee holds if consistent hashing selects a
majority of honest aggregators, which is likely when the majority of
stake is honest.

Biscotti achieves the above guarantees using polynomial commitments
(see Appendix~\ref{sec:crypto}) combined with verifiable secret
sharing~\cite{Shamir:1979} of individual updates. The update of
length $d$ is encoded as a $d$-degree polynomial, which can be
broken down into $n$ shares such that ($n = 2*(d+1)$).  These $n$
shares are distributed equally among $m$ aggregators. Since an update
can be reconstructed using $(d+1)$ shares, it would require $\frac{m}
{2}$ colluding aggregators to compromise the privacy of an individual
update. Therefore, given that any majority of aggregators is honest and
does not collude, the privacy of an individual peer update is
preserved. 

A peer with a verified update already possesses a
commitment $C = COMM(\Delta w_{i}(x))$ to its SGD update signed by a
majority of the verifiers from the previous step. To compute and
distribute its update shares among the aggregators, peer $i$ runs the
following secret sharing procedure: 

\begin{enumerate}
    \item  The peer computes the required set of secret shares $s_
    {m,i} = {z, \Delta w_{i}(z) | z \in Z}$ for aggregator $m$. 
    In order to ensure that an adversary does not provide shares from
    a poisoned update, the peer computes a set of associated witnesses
    $wit_{m,i} = \{COMM(\Psi_{z} (x)) | \Psi_{z}(x) = \frac{\Delta w
    (x) - \Delta w(z)}{x - z}\}$.
    These witnesses will allow the aggregator to verify that the secret
    share belongs to the update $\Delta w_i$ committed to in $C$. 
    It then sends $<C, s_{m,i} , wit_{m,i}>$ to each aggregator along with
    the signatures obtained in the verification stage.
    
    \item After receiving the above vector from peer $i$, the
    aggregator $m$ runs the following sequence of validations:

    \begin{enumerate}
      \item $m$ ensures that $C$ has passed the validation phase by
      verifying that it has the signature of the majority in the
      verification set.
      \item $m$ verifies that in each share $(z, \Delta w_{i}(z)) \in
      s_{m,i}$ $\Delta w_{i}(z)$ is the correct evaluation at $z$ of
      the polynomial committed to in $C$. (For details, see Appendix ~\ref{sec:crypto}) 

    \end{enumerate}
        
\end{enumerate}

Once every aggregator has received shares for the minimum number of updates
$u$ required for a block, each aggregator aggregates its individual
shares and shares the aggregate with all of the other aggregators. As soon
as a aggregator receives the aggregated $d+1$ shares from at least half of the
aggregators, it can compute the aggregate sum of the updates and create the
next block. The protocol to recover $\sum_{i=1}^{u} \Delta w_{i}$ is
as follows:
 
\begin{enumerate}     
  \item All m aggregators broadcast the sum of their accepted shares and
  witnesses $< \sum_{i=1}^{u} s_{m_i}, \sum_{j=1}^{u} wit_{m_j} >$    
  \item Each aggregator verifies the aggregated broadcast shares made by
  each of the other aggregators by checking the consistency of the
  aggregated shares and witnesses.     
  \item Given that $m$ obtains the shares from $\frac{m}{2}$ aggregators
  including itself, $m$ can interpolate the aggregated shares to
  determine the
  aggregated secret $\sum_{j=1}^{u} \Delta w_{j}$ 
\end{enumerate}

Once $m$ has figured out $\sum_{i=1}^{u} \Delta w_{i}$, it can create
a block with the updated global model.  All commitments to the updates
and the signature lists that contributed to the aggregate are
added to the block. The block is then disseminated in the network.
Any peer in the system can verify that all updates are verified by
looking at the signature list and homomorphically combine the
commitments to check that the update to the global model was computed
honestly (see ~\ref{subsec:blockchaindesign}). If any of these conditions 
are violated, the block is rejected.

\subsection{Blockchain consensus}

Because consistent hashing based subsets are globally observable by each peer, and
based only on the SHA-256 hash of the latest block in the chain,
ledger forks should rarely occur in Biscotti. For an update to be
included in the ledger at any iteration, the same
noising/verification/aggregation committees are used. Thus, race
conditions between aggregators will not cause forks in the ledger to occur
as frequently as in e.g., BitCoin~\cite{Nakamoto:2009}. 

When a peer observes a more recent ledger state through the gossip
protocol, it can catch up by verifying that the computation performed
is correct by running the consistent hashing protocol for the ledger state and by verifying
the signatures of the designated verifiers and aggregators for each
new block.

In Biscotti, each verification and aggregation step occurs only for a
specified duration. Any updates that are not successfully propagated
in this period of time are dropped: Biscotti does not append stale
updates to the model once competing blocks have been committed to the
ledger. This synchronous SGD model is acceptable for large scale ML
workloads which have been shown to be tolerant of bounded
asynchrony~\cite{Recht:2011}. However, these stale updates could be
leveraged in future iterations if their learning rate is decayed~\cite{Dean:2012}. We leave this optimization for future work.

%% file: implementation.tex
\section{Implementation}
\label{sec:impl}

We implemented Biscotti in 4,500 lines of Go 1.10 and 1,000 lines of
Python 2.7.12 and released it as an open source project{\footnote{\url{https://github.com/DistributedML/Biscotti}}}. We use Go to handle all networking and distributed systems
aspects of our design. We used PyTorch 0.4.1~\cite{Pytorch} to generate SGD
updates and noise during training. By building on the general-purpose
API in PyTorch, Biscotti can support any model that can be optimized
using SGD.  We use the \emph{go-python v1.0}~\cite{gopython}  library to
interface between Python and Go. Since go-python does not support Python $>=$ 2.7.12 therefore we were limited to using Python 2.7 for our implementation.

We use the \emph{kyber v.2}~\cite{Kyber} and \emph{CONIKS 0.1.1}~\cite{Coniks}
libraries to implement the cryptographic parts of our system. We use
CONIKS to implement our VRF function and kyber to implement the
commitment scheme and public/private key mechanisms. To bootstrap
clients with the noise commitments and public keys, we use an initial
genesis block. We used the bn256 curve API in kyber for generating our
commitments and public keys that form the basis of the aggregation
protocol and verifier signatures. For signing updates, we use the
Schnorr signature~\cite{Schnorr:1990} scheme instead of ECDSA because
multiple verifier Schnorr signatures can be aggregated together into
one signature~\cite{Maxwell:2018}. Therefore, our block size remains
constant as the verifier set grows.




%% file: eval.tex
\section{Evaluation}
\label{sec:eval}

We had several goals when designing the evaluation of our Biscotti
prototype. We wanted to demonstrate that (1) Biscotti is robust to
poisoning attacks, (2) Biscotti protects the privacy of an
individual client's data and (3) Biscotti is scalable, fault-tolerant
and can be used to train different ML models.

For experiments done in a distributed setting, we deployed Biscotti to
train an ML model across 20 Azure A4m v2 virtual machines, with 4 CPU
cores and 32 GB of RAM.  We deployed a varying number of peers in each
of the VMs. The VM's were spread across six locations: West US, East
US, Central India, Japan East, Australia East and Western Europe. We
measured the error of the global model against a test
set and ran each experiment for 100 iterations. To evaluate
Biscotti's defense mechanisms, we ran known inference and poisoning
attacks on federated learning~\cite{Melis:2018, Huang:2011} and
measured their effectiveness under various attack scenarios and
Biscotti parameters.  We also evaluated the performance implications
of our design by isolating specific components of our system and
varying the committee sizes with different numbers of peers. For all our experiments, we deployed Biscotti with the parameter values in
Table~\ref{tab:defaultparameters} unless stated otherwise. 

We evaluated Biscotti with logistic regression and softmax
classifiers on the Credit Card and MNIST dataset respectively. The softmax classifier contains a one layer neural network with a binary cross entropy loss at the end. However, due to the general-purpose PyTorch API, we claim
that Biscotti can generalize to models of arbitrary size and
complexity, as long as they can be optimized with SGD and can be
stored in our block structure. We evaluate logistic regression with a
credit card fraud dataset from the UC Irvine data
repository~\cite{Dua:2017}, which uses an individual's financial and
personal information to predict whether or not they will default on
their next credit card payment. We evaluate softmax with the canonical
MNIST~\cite{Lecun:1998} dataset, a task that involves predicting a
digit based on its image. 

The MNIST and Credit Card datasets have 60,000 and 21000 training examples respectively. We used 5-fold cross validation on the training set to decide on hyperparameters of batch size, momentum and learning rate indicated in Table ~\ref{tab:evalhyperparameters}. The MNIST/Credit Card models were tested using a separately held-out test set of 10,000 and 9000 examples respectively.  For all experiments, the  training dataset was divided equally among the peers unless stated otherwise. As a result, each client possesses 600 training examples for MNIST and 210 examples for the credit card dataset. The size of dataset/training examples per peer is small for our experimental setting and might not be representative of large-scale datasets.     

In summary, Table~\ref{tab:evalmodels} and ~\ref{tab:evalhyperparameters} show
the datasets, types of model, number of training examples, the 
number of parameters $d$ in the models and the hyperparameters that were used in our experiments. 

In local training we were able to train a model on the
Credit Card and MNIST datasets with accuracy of 98\% and 92\%,
respectively.

\begin{table}[t]
{\small
\centering
\begin{tabular}{r|c|c|c}
 \textbf{Dataset} & \textbf{Model Type} & \textbf{Train/Test Examples}  & \textbf{Params ($d$)} \\
 \hline
 \textbf{Credit Card} & LogReg & 21000/9000 & 25 \\
 \hline
 \textbf{MNIST} & Softmax & 60000/10000 & 7850 \\ 
 \hline
\end{tabular} 
}
\caption{The dataset/model types used in the experiments}
\label{tab:evalmodels}
\end{table}

\begin{table}[t]
{\small
\centering
\begin{tabular}{r|c|c|c}
 \textbf{Model/Dataset} & \textbf{Batch Size} & \textbf{Learning Rate} & \textbf{Momentum} \\
 \hline
 \textbf{Credit Card/LogReg} & 10 & 0.01 & 0 \\
 \hline
 \textbf{MNIST/Softmax} & 10 & 0.001 & 0.75 \\
 \hline
\end{tabular} 
}
\caption{The hyperparameters used in the experiments}
\label{tab:evalhyperparameters}
\end{table}



\begin{table}[t]
{\small
\centering
\begin{tabular}{p{5cm}|l}
 \textbf{Parameter} & \textbf{Default Value} \\
 \hline
 Privacy budget ($\varepsilon$) & 2 \\
  \hline
 Delta ($\delta$) & $10^{-5}$ \\
 \hline
 Number of peers & 100 \\
 \hline
 Number of noisers & 2 \\
 \hline
 Number of verifiers & 3 \\
 \hline
 Number of aggregators & 3 \\
 \hline
 Number of samples needed for Multi-KRUM (R) & 70 \\
 \hline
 Adversary upper bound ($f < \frac{R-2}{2}$) & 33 \\
 \hline
 Number of updates/block ($u=\frac{R}{2}$)  & 35 \\
 \hline
 Initial stake & Uniform, 10 each \\
 \hline
 Stake update & Linear, + 5 \\ 
\end{tabular} 
}
\caption{The default parameters used for all our experiments,
unless stated otherwise.}
\label{tab:defaultparameters}
\end{table}

\subsection{Minimum committee size to prevent collusion}
\label{eval:vrfsecurity}

In Biscotti, the verification and aggregation stages involve committees that use a majority voting scheme to reach consensus. By making these committee sizes large enough, we can prevent an adversary controlling a certain fraction of the stake from acting maliciously. An adversary having the majority vote can act maliciously by accepting poisoned updates or recovering an individual peer's updates during aggregation. In this section, we carry out an analysis of the least committee size needed such that the probability of an adversary having the majority is below a threshold.

Using the consistent hashing protocol, the probability of a peer being selected is proportional to their stake. Hence, the probability $p$ of an adversary having the majority in a committee size $k$ can be calculated by:
  
\begin{align*}
  p = \sum_{(i=\frac{k}{2} + 1)}^{k} {k\choose{i}}s^{i}(1-s)^{k-i} 
\end{align*}

where $s$ is the fraction of stake controlled by the adversary.

By assuming that $p$ follows a binomial distribution, we obtain a loose upper bound for an adversary controlling the majority in Biscotti. A binomial distribution assumes sampling peers with replacement allowing a peer to be elected more than once in the committee. Since Biscotti limits a peer to only one vote in the committee, the actual probability of the adversary controlling the majority in Biscotti is less than $p$. 

Since $p$ is an upper bound, we can safely use it to calculate the smallest committee size that bounds $p$ below a threshold ($t$). To obtain the minimum committee size for $p$, we use a brute force approach and try out different committee sizes and pick the least size that causes $p$ to fall below the threshold.

\begin{figure}[t]
  \centering
  \includegraphics[width=\linewidth]{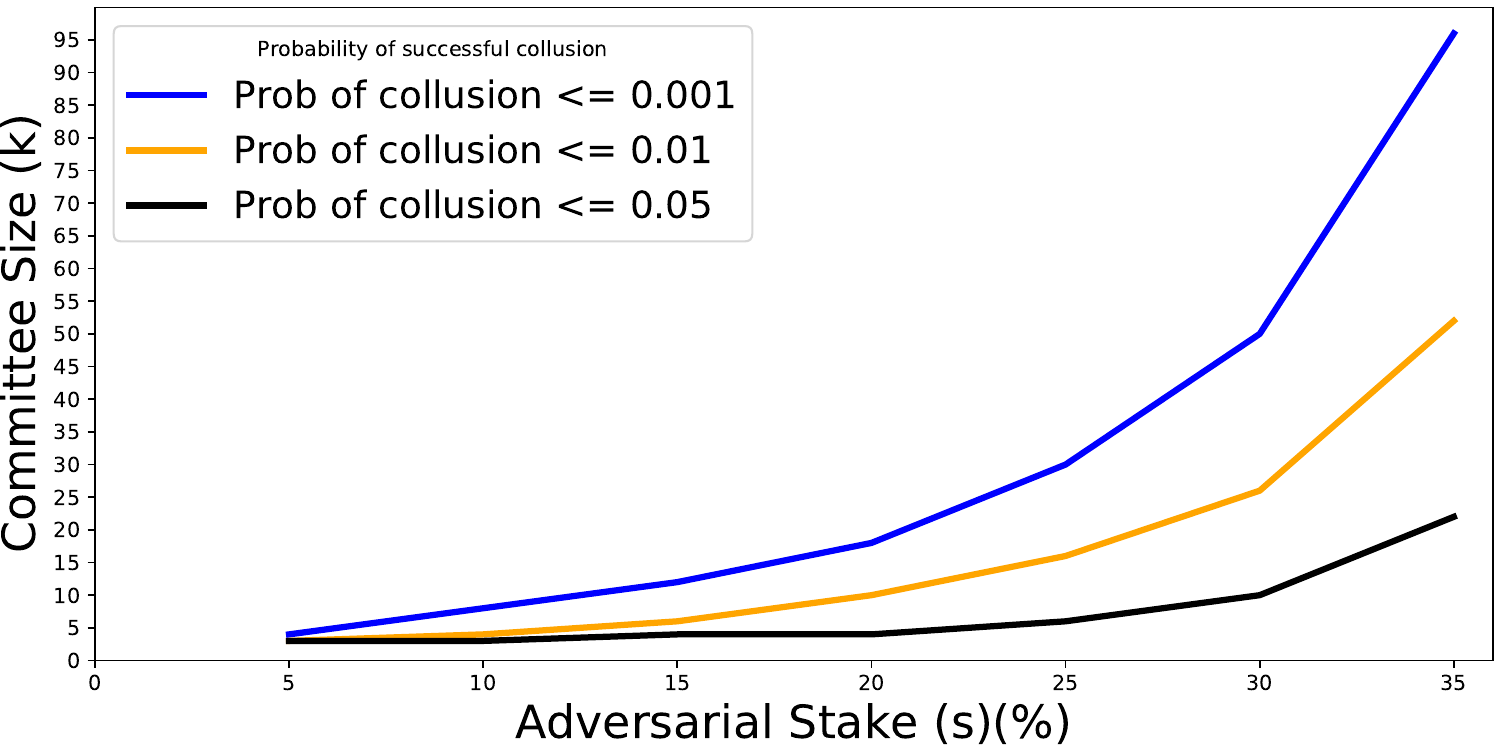}
  \caption{Size of committee needed such that the probability of an adversary successfully colluding is below a threshold.}
  \label{fig:evalvrfsecurity}
\end{figure}

Figure~\ref{fig:evalvrfsecurity} plots the minimum committee size needed against adversarial stake for probability thresholds of 0.01, 0.05 and 0.001 respectively. The minimum committee size is independent of the number of nodes and grows exponentially with adversarial stake in the system. Since our experimental evaluation is limited to training for a 100 rounds, the probability threshold of an adversary controlling the majority $t$ needs to be less than 0.01. For this threshold, a committee size of 26 protects against an adversary controlling 30\% of the stake in the system.

\subsection{Tolerating poisoning attacks}
\label{eval:poisoning}

In this section, we evaluate Biscotti's performance when we subject
the system to a label flipping poisoning attack as in Huang et.al~\cite{Huang:2011}. We investigate the different parameter settings required for Biscotti to successfully defend against a poisoning attack and then evaluate how well it performs under attack from 30\% malicious nodes compared to a Federated Learning baseline. 

\begin{figure}[t]
  \centering
  \includegraphics[width=\linewidth]{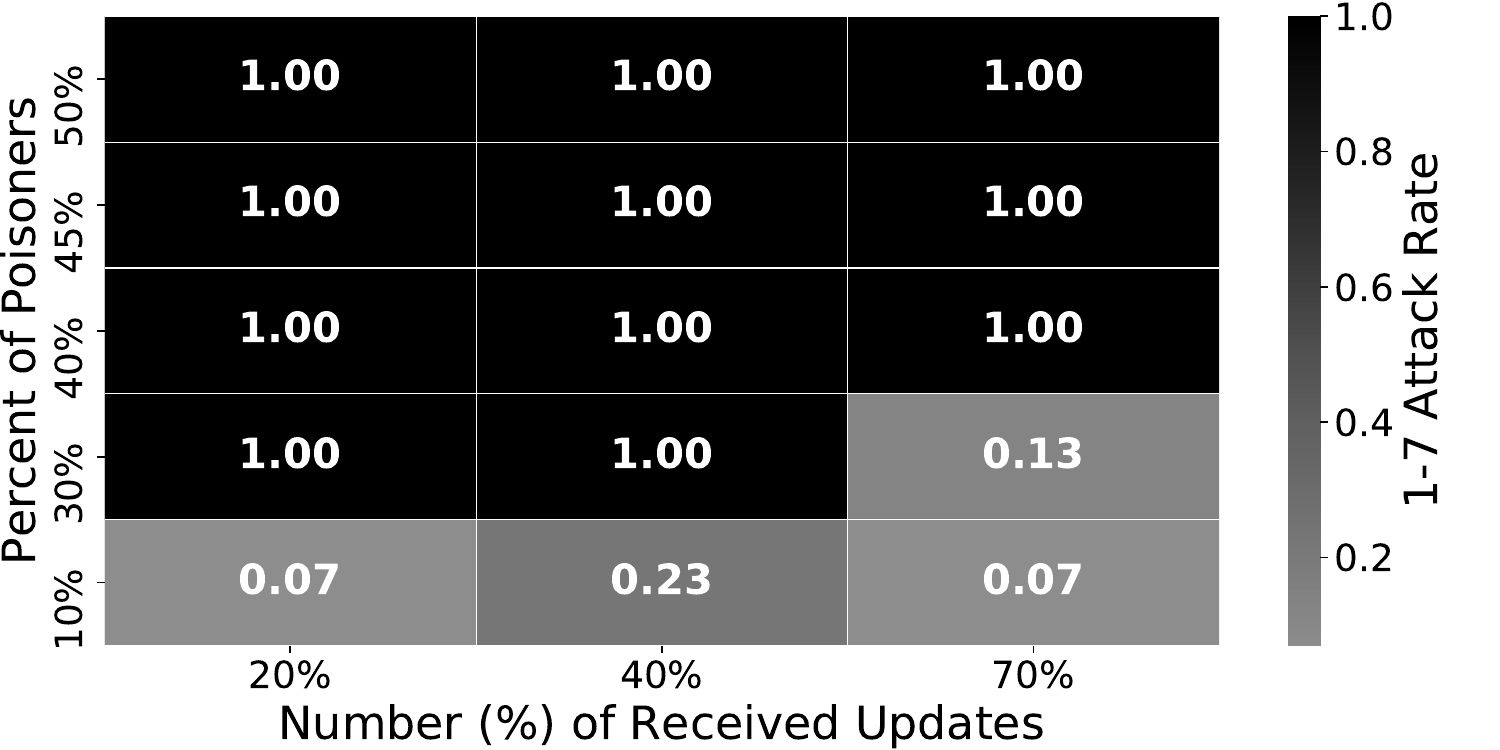}
  \caption{Evaluating the effect of the number of sampled updates each round on KRUM's performance.}
  \label{fig:evalpoisonheatmap}
\end{figure}

\noindent \textbf{Attack Rate vs Received Updates.}
Biscotti requires each peer in the verification committee to collect a
sufficient sample of updates before running Multi-KRUM.  We evaluated
the effect of collecting varying percentages of total updates in each
round on the effectiveness of Multi-KRUM with different poisoning
rates for the MNIST dataset. To ensure uniformity and to eliminate latency effects in the
collection of updates, in these experiments the verifiers waited for
updates from all peers that are not assigned to a committee and
then randomly sampled a specified number of updates. In addition, we
also ensured that all verifiers deterministically sampled the
\emph{same} set of updates by using the last block hash as the random
seed. This allows us to investigate how well Multi-KRUM performs in a
decentralized ML setting like Biscotti unlike the original Multi-KRUM
paper~\cite{Blanchard:2017} in which updates from \emph{all} clients
are collected before running Multi-KRUM.

To evaluate the success of an attack, we define \emph{attack rate} as
the fraction of target labels that are incorrectly predicted. An
\emph{attack rate} of 1 specifies a successful attack while a value
close to zero indicates an unsuccessful one.

Figure~\ref{fig:evalpoisonheatmap} show the results. As the percentage
of poisoners in the system increases, a higher fraction of updates
need to be collected for Multi-KRUM to be effective (to achieve a low
attack rate). A large sample ensures that the poisoners do not gain
majority in the set being fed to Multi-KRUM, otherwise Multi-KRUM
cannot prevent poisoned updates from leaking into the model. The results show that each round updates need to be selected from 70\% of the peers to prevent poisoning from 30\% of the nodes. 

\begin{figure}[t]
  \centering
  \includegraphics[width=\linewidth]{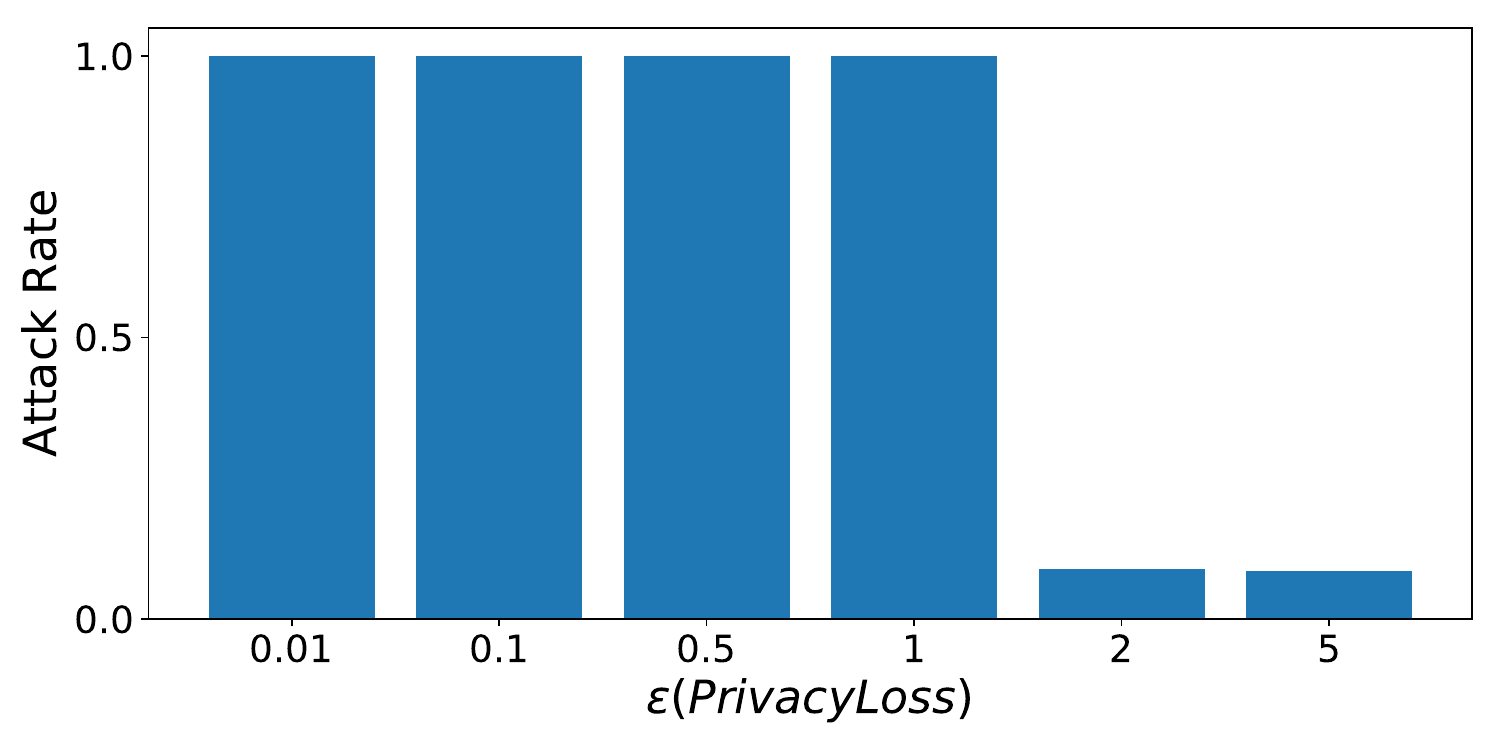}
  \caption{KRUM's performance in defending against a 30\% attack on
    the MNIST dataset for different settings of $\epsilon$ noise.}
  \label{fig:evalnoisekrum}
\end{figure}

\noindent \textbf{Attack Rate vs Noise.}
To ensure that updates are kept private in the verification stage,
differentially private noise is added to each update before it is sent
to the verifier. This noise is parametrized by the $\epsilon$ and
$\delta$ parameters. The $\delta$ parameter indicates the probability
with which plain $\epsilon$-differential privacy is broken and is
ideally set to a value lower than $1/|d|$ where d is the dimension of
the dataset. Hence, we set $\delta$ to be $10^{-5}$ in all our
experiments.  $\epsilon$ represents privacy-loss and a lower value of
$\epsilon$ indicates that more noise is added to the updates. We
investigate the effect of the $\epsilon$ value on the performance of
Multi-KRUM with 30\% poisoners in a 100-node deployment with 70
received updates in each round on the MNIST dataset.
Figure~\ref{fig:evalnoisekrum} shows that KRUM loses its effectiveness
at values of $\epsilon \leq 1$ but performs well on values of
$\epsilon \geq 2$.


\begin{figure}[t]
  \centering
  \includegraphics[width=\linewidth]{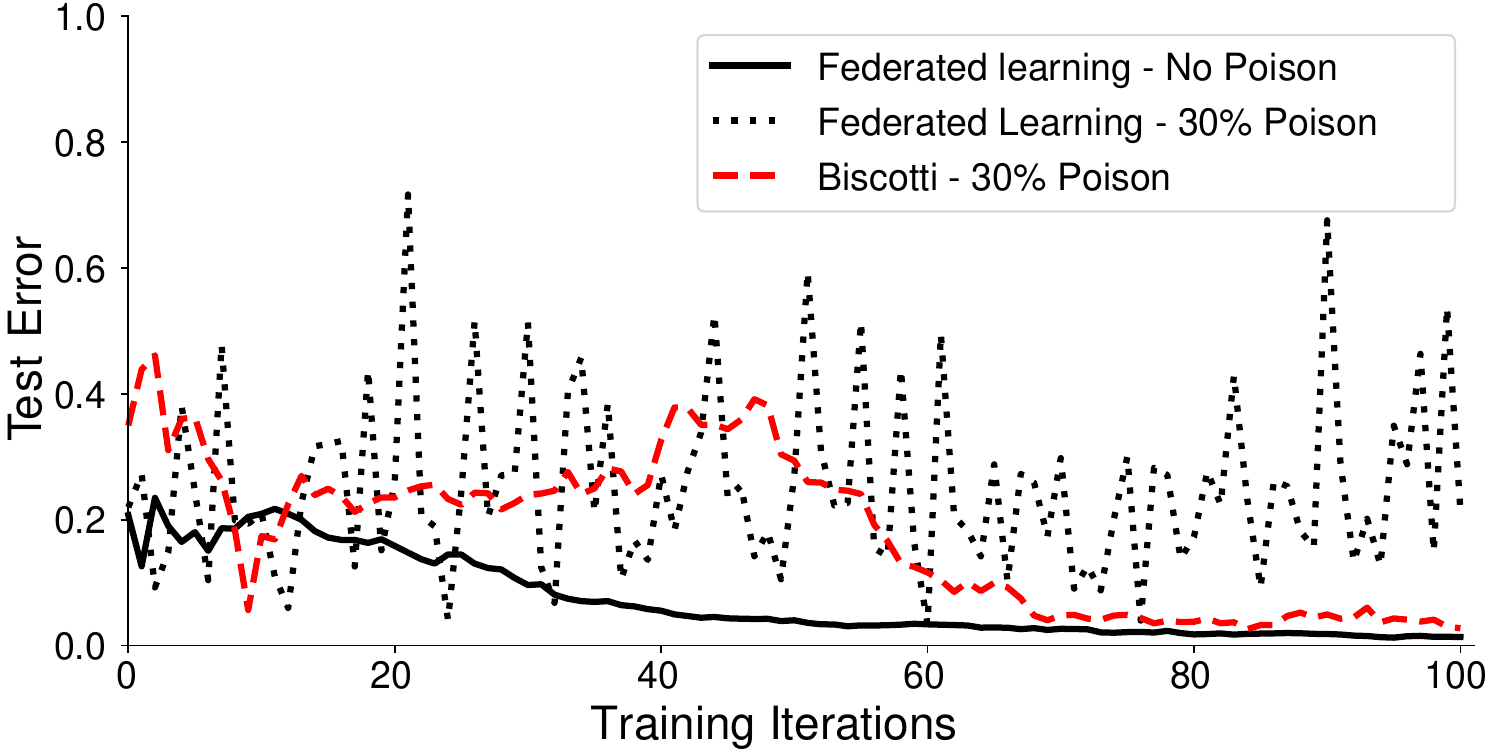}
  \caption{Federated learning and Biscotti's test error on the CreditCard dataset with 30\% of poisoners.}
  \label{fig:evalpoisoning30credit}
\end{figure}

\begin{figure}[t]
  \centering
  \includegraphics[width=\linewidth]{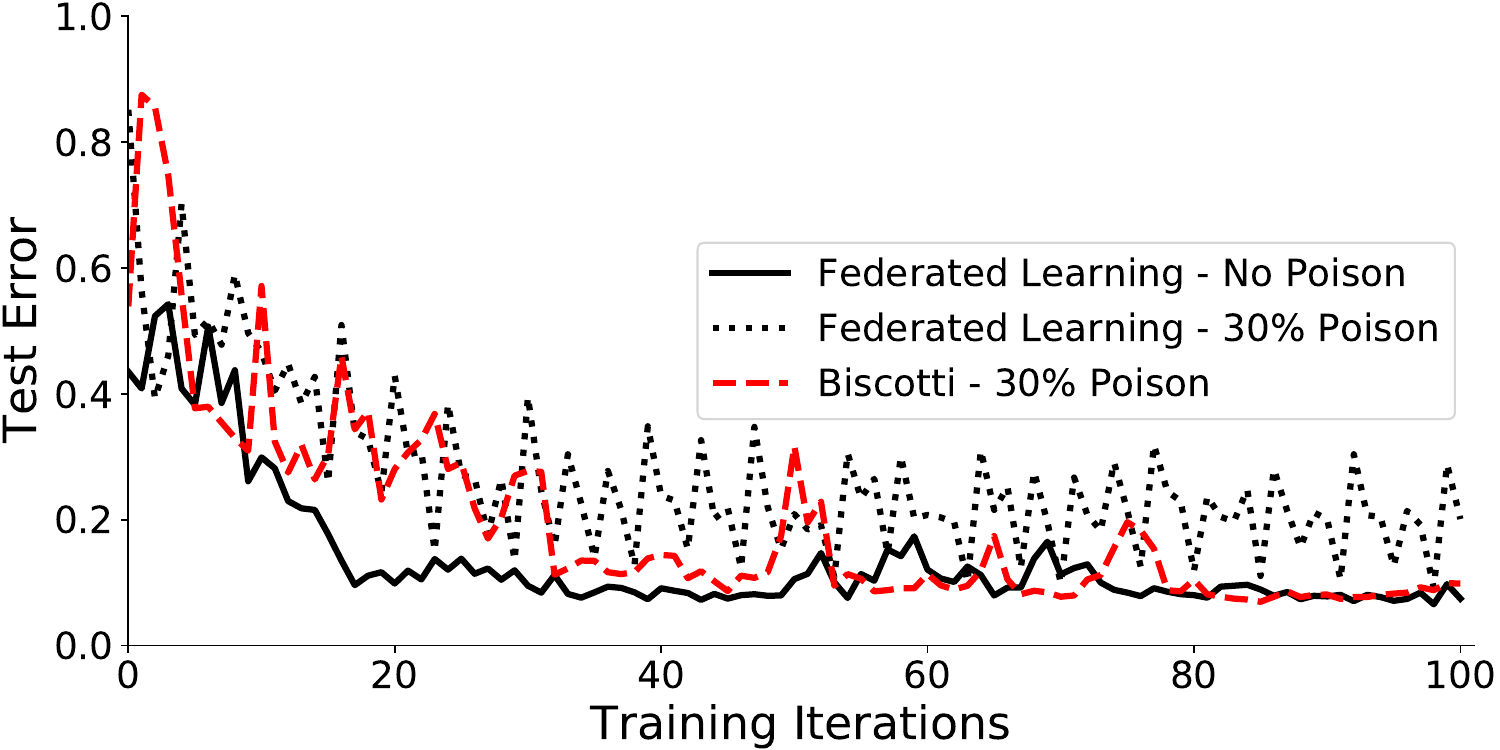}
  \caption{Federated learning and Biscotti's test error on the MNIST dataset with 30\% of poisoners.}
  \label{fig:evalpoisoning30mnist}

\end{figure}

\begin{figure}[t]
  \centering
  \includegraphics[width=\linewidth]{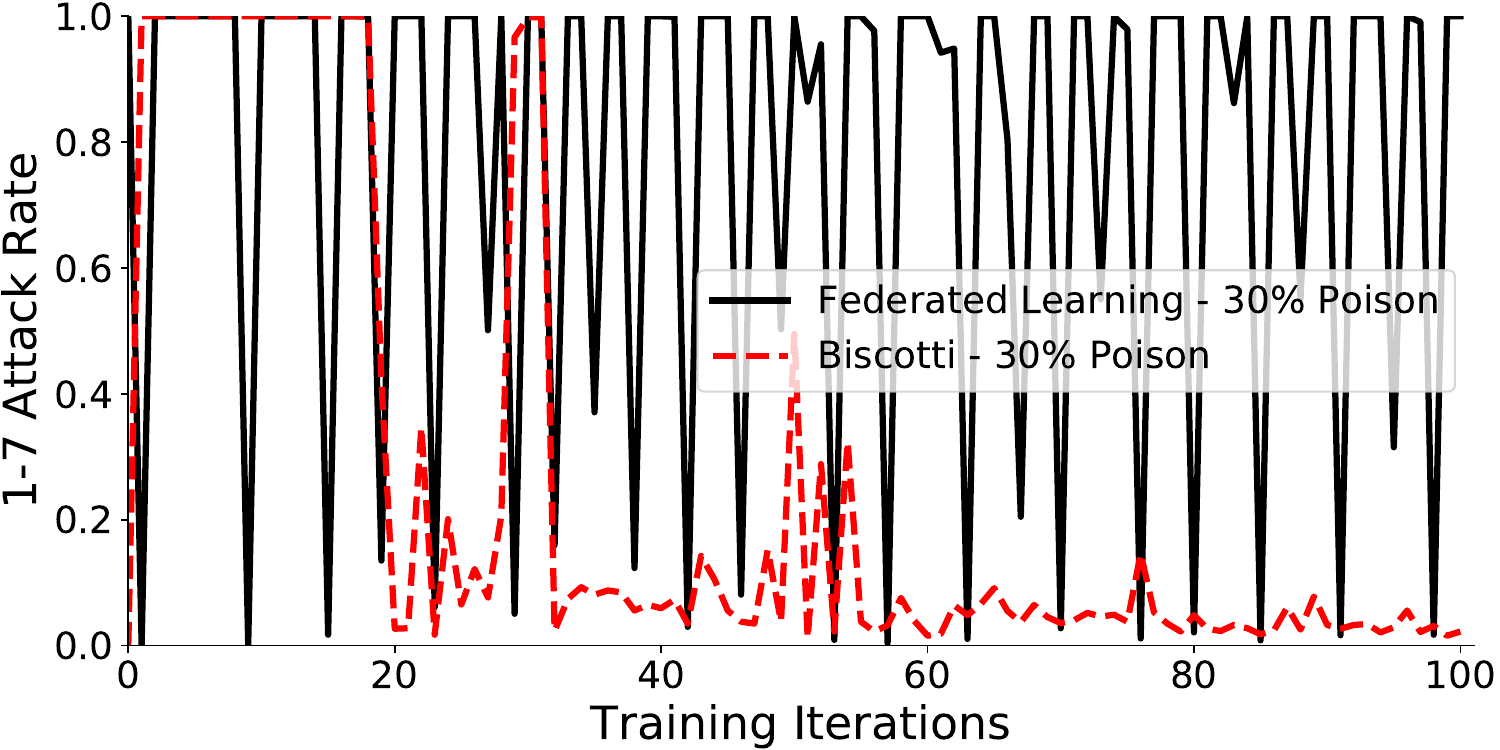}
  \caption{Attack rate comparison of federated learning and Biscotti's on the MNIST dataset with 30\% of poisoners.}
  \label{fig:evalpoisoning30attack}
\end{figure}

\begin{figure}[t]
  \centering
  \includegraphics[width=\linewidth]{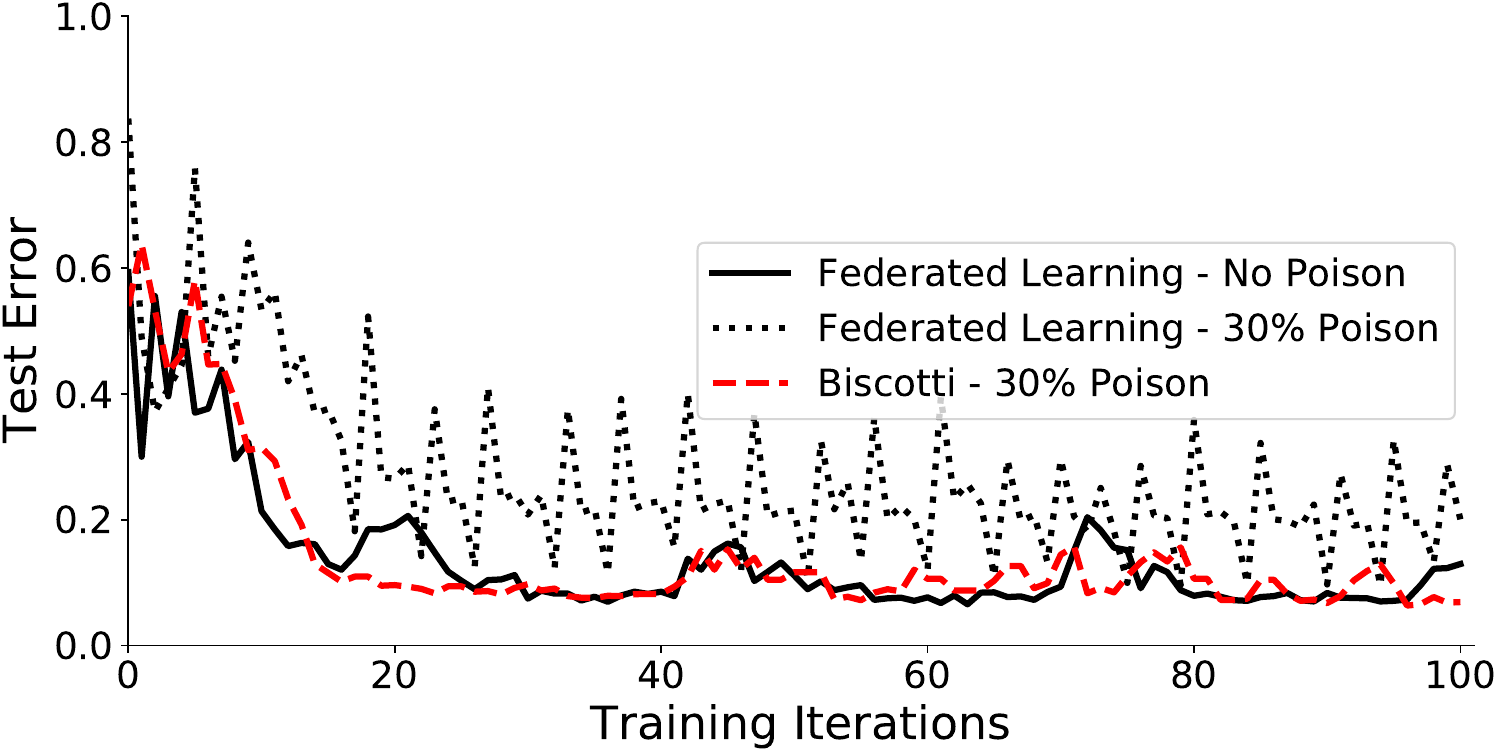}
  \caption{Federated learning and Biscotti's test error on the MNIST dataset with 30\% of poisoners with larger committee size.}
  \label{fig:evalpoisoningcommittee}
\end{figure}

\begin{figure}[t]
  \centering
  \includegraphics[width=\linewidth]{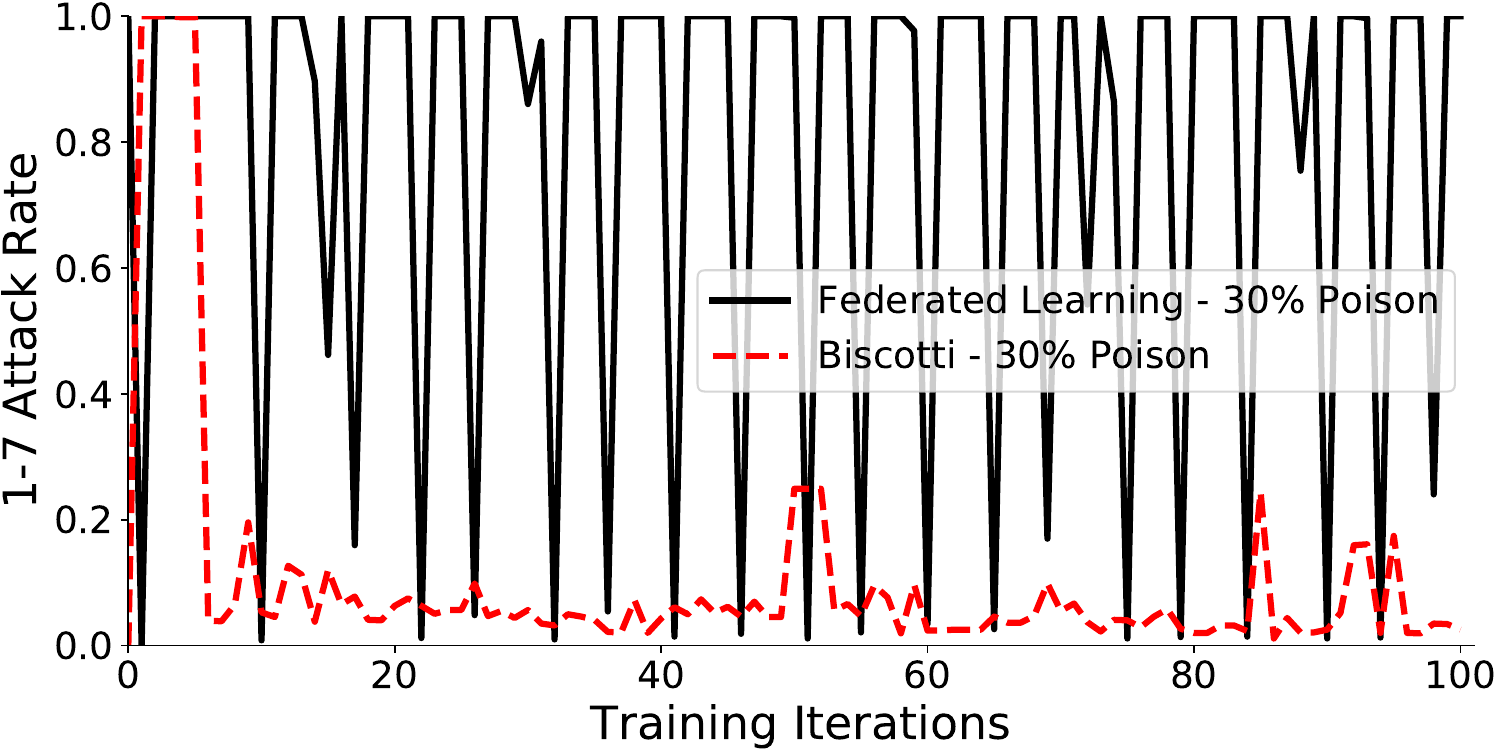}
  \caption{Attack rate comparison of federated learning and Biscotti's on the MNIST dataset with 30\% of poisoners with larger committee size.}
  \label{fig:evalpoisoningcommitteeattack}
\end{figure}


%
\noindent \textbf{Biscotti vs Federated Learning Baseline.}
We deployed Biscotti and federated learning and subjected both systems
to a poisoning attack while training on the Credit Card and MNIST
datasets. Using the parameters from the above experiments, Biscotti sampled 70\% of updates and used a value of 2 for the privacy budget $\epsilon$. 

We introduced 30\% of the peers into the system with the same
malicious dataset: for credit card they labeled all defaults as
non-defaults, and for MNIST these peers labeled all 1s as
7s. Figure~\ref{fig:evalpoisoning30credit}
and~\ref{fig:evalpoisoning30mnist} shows the test error for both
datasets as compared to federated learning. The results show that for
both datasets the poisoned federated learning deployment struggled to
converge. By contrast, Biscotti performed as well as the baseline
un-poisoned federated learning deployment on the test set.


 We show the attack rate in Biscotti and
Federated Learning for MNIST in Figure~\ref{fig:evalpoisoning30attack}. For the credit card dataset, the
attack rate is the same as the test error since there is only one
class in the dataset. The figure illustrates that in federated learning, since no defense
mechanisms exist, a poisoning attack creates a tug of war between
honest and malicious peers. On the other hand, in Biscotti the
attack rate settles after some initial jostling. 

The jostling results in Biscotti taking a larger number of iterations to converge than un-poisoned federated learning. The convergence is slower for Biscotti because a small verification committee of 3 peers was used for this experiment. A smaller committee size allowed the poisoning peers to control a majority in the committee frequently and accept updates from fellow malicious peers. Biscotti finally converged because the honest peers gained influence over time thereby reducing the chance of malicious peers from taking control of the majority vote each round. In this experiment, the
stake of the honest clients went from 70\% to 87\%.

To demonstrate the effect of the committee size on convergence under attack, we run the same experiment for the MNIST dataset with an increased committee size of 26 peers. As shown in Section \ref{eval:vrfsecurity}, a committee size of 26 provides protection against an adversary controlling 30\% of the stake in the system. Since peers part of committees do not contribute updates in that particular round, 100 nodes is not enough to collect 70\% of updates needed to protect against Multi-KRUM. Therefore, for this experiment we increased the number of nodes to 200. The results in Figure ~\ref{fig:evalpoisoningcommitteeattack} shows that Biscotti converges in the same number of iterations as unpoisoned Federated Learning with an increased committee size. Figure ~\ref{fig:evalpoisoningcommitteeattack} shows that the attack rate stays less than 24.9\% after the first 5 iteration because the poisoners were unable to get their updates accepted during the training process.




\subsection{Privacy evaluation}
\label{eval:privacy}

In this section, we evaluate the privacy provided by secure
aggregation in Biscotti. We subject Biscotti to an information leakage
attack~\cite{Melis:2018} and demonstrate that the effectiveness of
this attack decreases with the number of securely aggregated updates
in a block. We also show that the probability of a successful
collusion attack to recover an individual client's private gradient
decreases as the size of the committees grows.

\begin{figure*}[t]
  \centering
  \includegraphics[width=\linewidth]{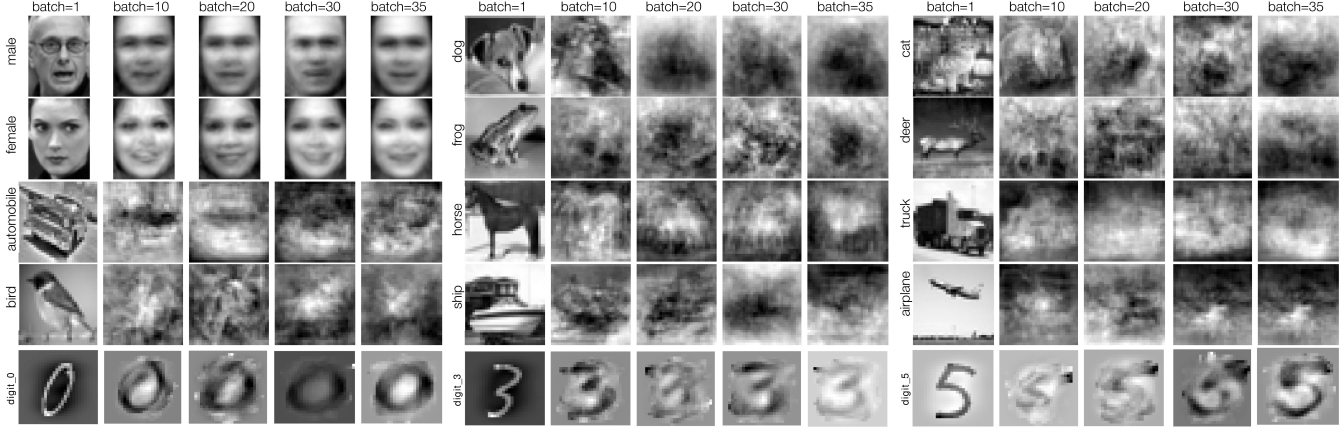}
  \caption{The results of an information leakage attack on
    different number of aggregated gradients for different classes.}
  \label{fig:evalbatching}
\end{figure*}

\noindent \textbf{Information leakage from aggregated gradients.}
We subject the aggregated gradients from several different datasets to
the gradient-based information leakage attack described
in~\cite{Melis:2018}. We invert the aggregated gradient knowing that
the gradient for the weights associated with each class in the fully
connected softmax layer is directly proportional to the input
features. To infer the original features, we take the gradients from a
single class and invert them with respect to all the classes in the CIFAR-10 (10 classes of objects/animals) and LFW datasets(2 classes male/female). We also invert the gradients for three classes (0,3,5)  on the MNIST dataset. We visualize the gradient in grayscale after reshaping
to the original feature dimensions in
Figure~\ref{fig:evalbatching}. The aggregated gradient will have data
sampled from a mixture of classes including the target class. Our
results show that having a larger number of participants in the
aggregate decreases the impact of this attack. As shown in Figure~\ref{fig:evalbatching}, as the number of aggregates batched together increases, it becomes harder to distinguish the individual training examples. 

By default Biscotti aggregates/batches 35
updates. Figure~\ref{fig:evalbatching} illustrates how the individual
class examples from the inverted images are difficult to determine. 
It might be easy to infer what a class looks like from the inversions. But if the class does not represent an individual's training set, secure aggregation provides privacy. 
For example, in LFW we get an image that represents what a male/female looks
like but we do not gain any information about the individual training
examples. Hence, the privacy gained is dependent on how close the
training examples are to the class representative.

\begin{figure}[t]
  \centering
  \includegraphics[width=\linewidth]{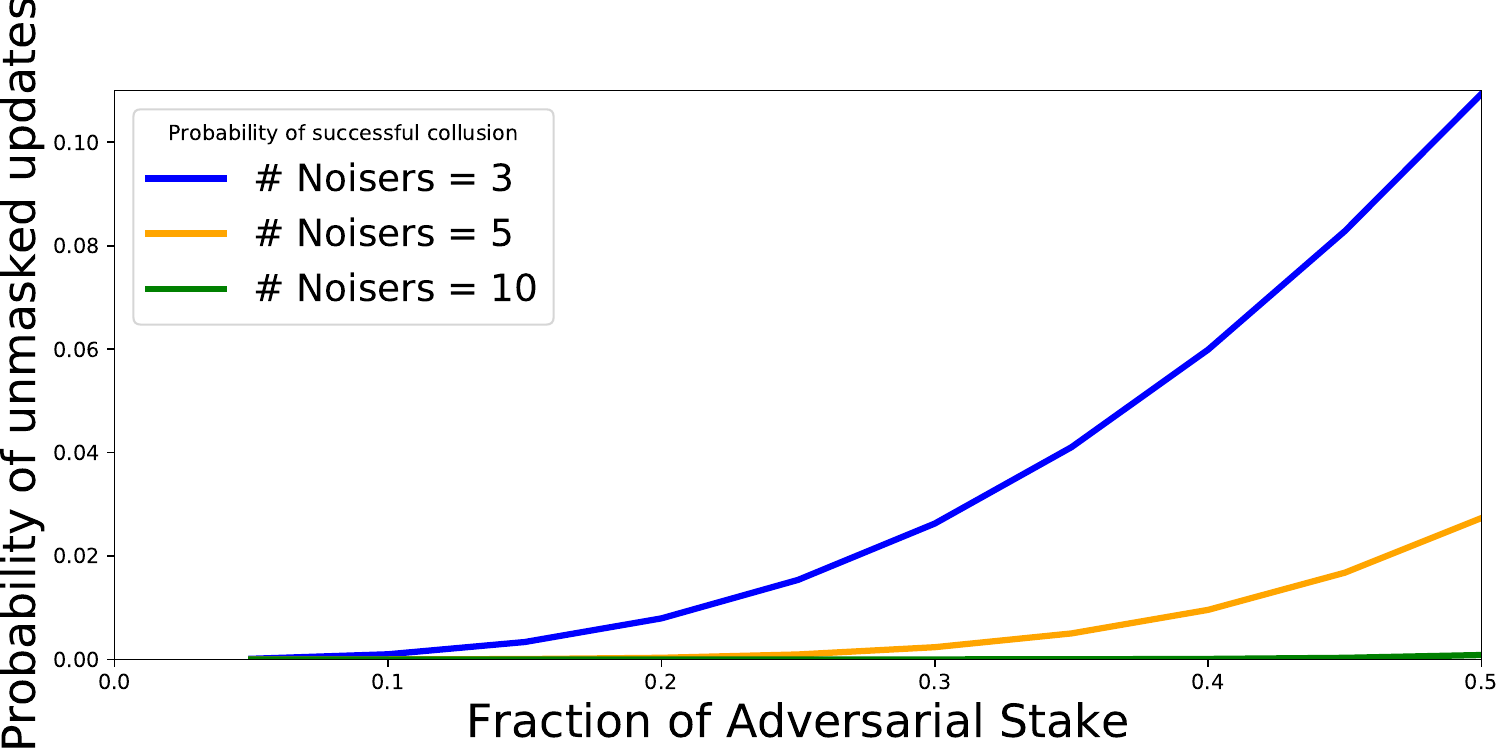}
  \caption{Probability of a successful collusion attack to recover an individual client's gradient. 
  }
  \label{fig:evalnoiseattack}
\end{figure}

\noindent \textbf{Recovering a client's individual gradient.}

We also evaluate a proposed attack on the noising protocol, which aims to de-anonymize peer gradients. This attack is performed when a
verifier colludes with several malicious peers. When bootstrapping
the system, the malicious peers pre-commit noise that sums to 0.
As a result, when a noising committee selects these noise elements for
verification, the masked gradient is not actually masked with any
$\varepsilon$ noise, allowing a malicious peer to perform an
information leakage attack on the victim.


We evaluated the effectiveness of this attack by varying the proportion of malicious stake in the system and calculating the probability of the adversary being able to unmask the updates for various sizes of the noising committee. A malicious peer can unmask an update if it controls all the noisers for a peer and has atleast one verifier in the verification committee.  Figure~\ref{fig:evalnoiseattack} shows the probability of a privacy violation as the proportion of adversarial stake increases for noising committee sizes of 3, 5 and 10 respectively.




When the number of noisers for an iteration is 3, an adversary
needs at least 15\% of stake to successfully unmask an SGD
update. This trend continues when 5 noisers are used: over 30\% of the
stake must be malicious. When the number of noisers is 10 (which has
minimal additional overhead according to Figure~\ref{fig:evalnva}),
privacy violations do not occur even with 50\% of malicious stake. By
using a stake-based consistent hashing to select noising clients, Biscotti
prevents adversaries from performing information leakage attacks on
other clients unless their proportion of stake in the system is
overwhelmingly large.

\subsection{Performance, scalability and fault tolerance}

In this section, we evaluate the overhead of each stage in Biscotti and investigate the effect on the overhead as the number of peers increase. In addition to this, we also measure the effect on Biscotti's performance as we scale the size of different committees. We compare the performance of Biscotti on federated learning. Finally, we see if churn has any effect on Biscotti's convergence.

\begin{figure}[t]
  \centering
  \includegraphics[width=\linewidth]{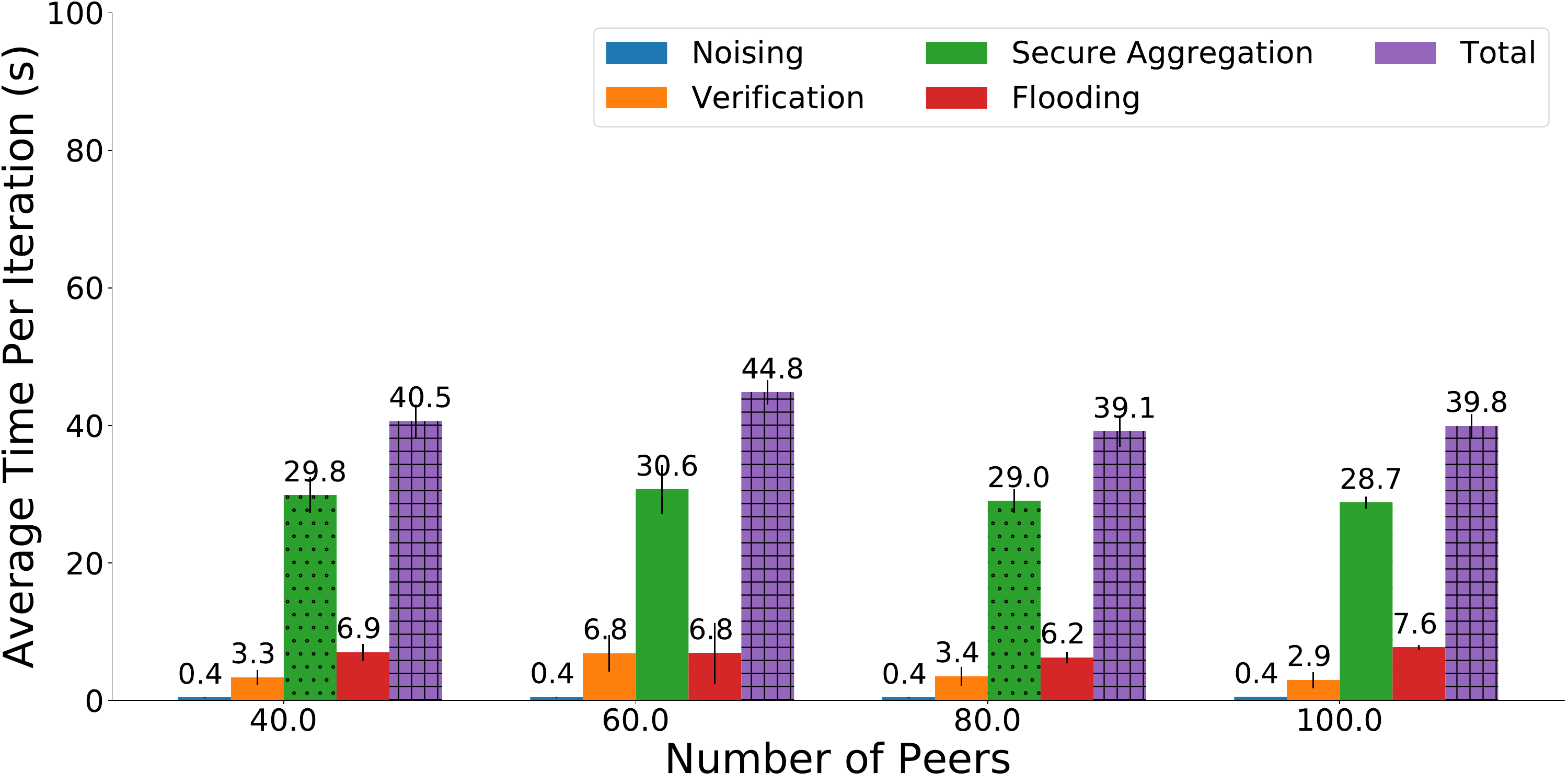}
  \caption{Breakdown of time spent in different mechanisms in Biscotti
    (Figure~\ref{fig:design}) in deployments with varying number of
    peers.
    %
}
  \label{fig:evalcost}
\end{figure}

\noindent \textbf{Performance cost break-down.}  In breaking down the
overhead in Biscotti, we deployed Biscotti over a varying number peers in training on MNIST. We captured the total amount of time spent in each of the major phases of our algorithm in Figure~\ref{fig:design}: collecting the noise from each of the noising clients (steps
\getnoise~and \noisyupdate), executing verification via Multi-KRUM and collecting the signatures (steps \krum ~and \signedupdate) and
securely aggregating the SGD update (steps \shamir~and
\secagg). Figure~\ref{fig:evalcost} shows the breakdown of the average cost per iteration for each stage under a deployment of 40, 60, 80 and 100 nodes over 3 runs. During this experiment, the committee sizes were kept constant to the default values in Table ~\ref{tab:defaultparameters}.

The results show that the cost of each stage is almost constant as we
vary the number of peers in the system. Biscotti spends most of its
time in the aggregation phase since it requires the aggregators to
collect secret shares of all the accepted updates and share the
aggregated shares with each other to generate a block. The noising
phase is the fastest since it only involves making a round trip to
each of the noisers while the verification stage involves collecting a
predefined percentage (70\%) of updates to run Multi-KRUM in a asynchronous
manner from all the nodes. The time per iteration also stays fairly
constant as the number of nodes in the system increase.



\begin{figure}[t]
  \centering
  \includegraphics[width=\linewidth]{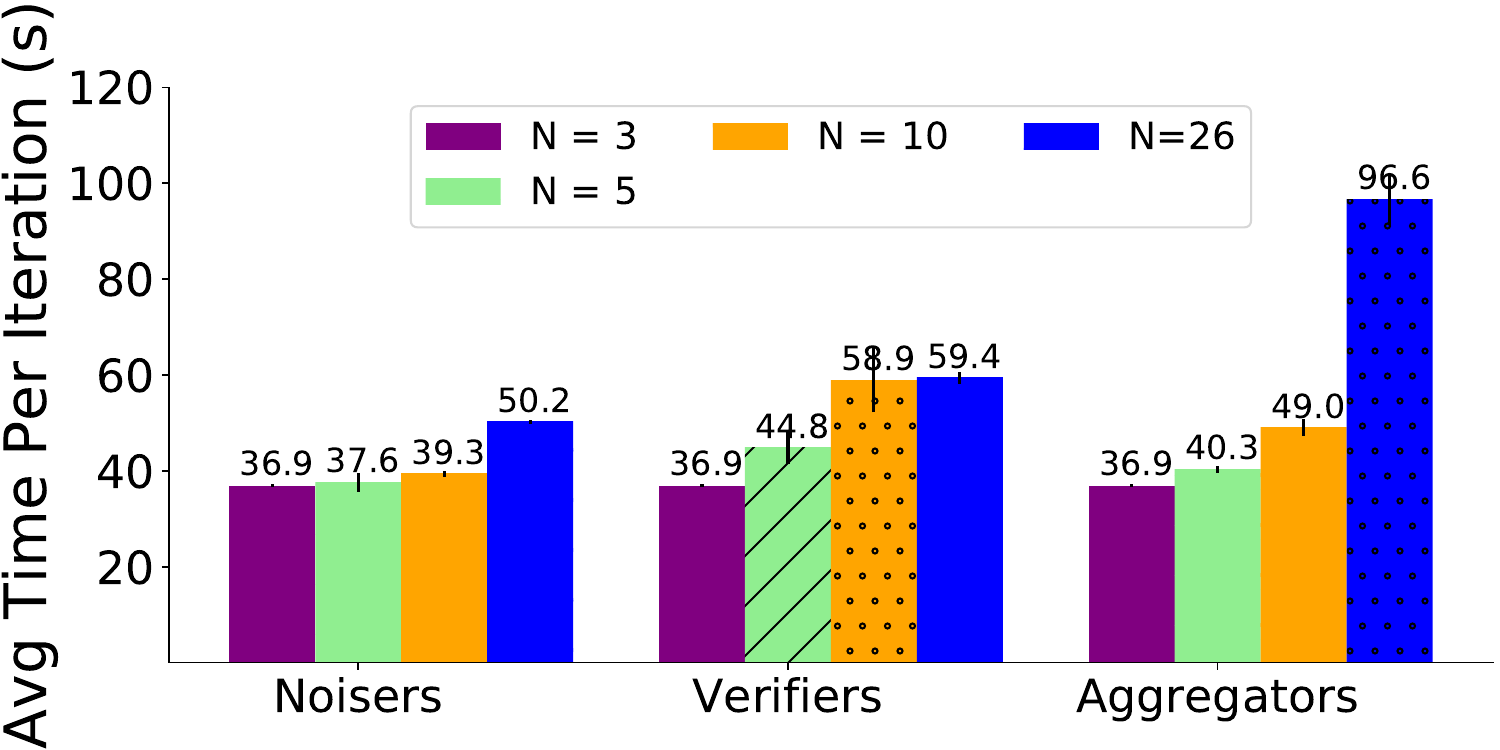}
  \caption{The average amount of time taken per iteration with an
  increasing number of noisers, verifiers, or aggregators.}
  \label{fig:evalnva}
\end{figure}

\noindent \textbf{Scaling up committee sizes in Biscotti.}
We evaluate Biscotti's performance as we change the size of the
noiser/verifier/aggregator sets.
For this we deploy Biscotti on Azure with the MNIST dataset with a
fixed size of 100 peers, and only vary the number of noisers needed for
each SGD update, number of verifiers used for each verification
committee, and the number of aggregators used in secure
aggregation. Each time one of these sizes was changed, the change was
performed in isolation; the rest of the committees used a set of 3. Figure~\ref{fig:evalnva} plots
the average time taken per iteration over 3 runs of the system.

The results show that increasing the number of noisers, verifiers and
aggregator sets increases the time per iteration. The iteration time increases slightly with noisers because it requires contacting additional peers for the noise. Increasing the aggregators leads to a larger overhead because the secret shares are shared with more aggregators and recovering the aggregate requires coordination among more peers. Lastly, a large number of verifiers results in frequent timeouts in the mining stage. Verifiers wait for the first 70  updates and select 37 updates while miners wait for shares from the first 35 updates before initiating the aggregate recovery process. With an increased verifier set, the size of the intersection of updates accepted by majority verifiers falls frequently below 35 because each of them run Multi-KRUM on a different set of updates. Hence, the aggregators wait until a timeout for 35 updates when the actual number of updates accepted during the round is less. Since the timeout is a constant value of 90 seconds, the verifier overhead does not increase significantly when increasing the verifier set from 10 to 26. However, this increased verifier overhead could be lowered by decreasing the number of updates that aggregators wait for before starting the coordination.

\begin{figure}[t]
  \centering
  \includegraphics[width=\linewidth]{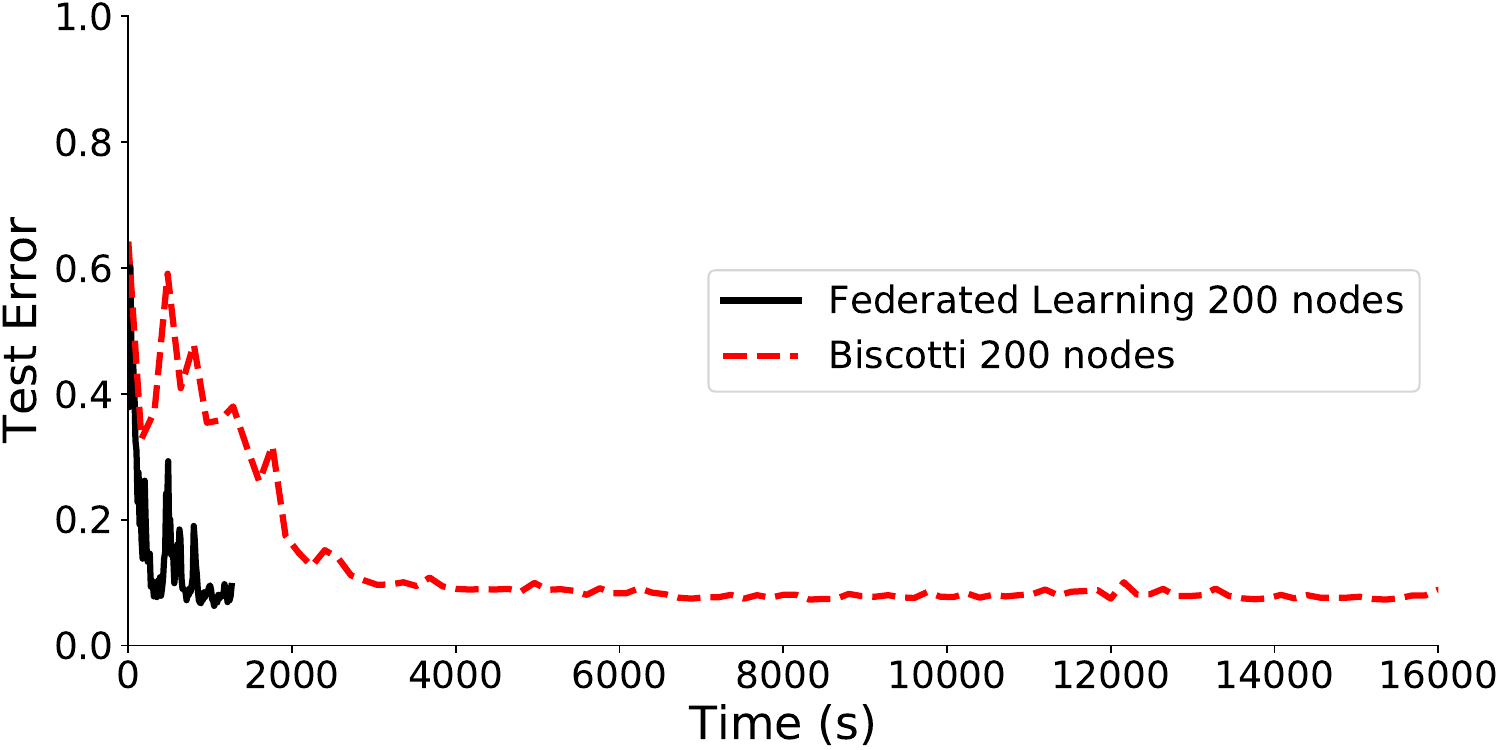}
  \caption{Comparing the time to convergence of Biscotti to a
  federated learning baseline.
  }
  \label{fig:evalconvtime}
\end{figure}

\begin{figure}[t]
  \centering
  \includegraphics[width=\linewidth]{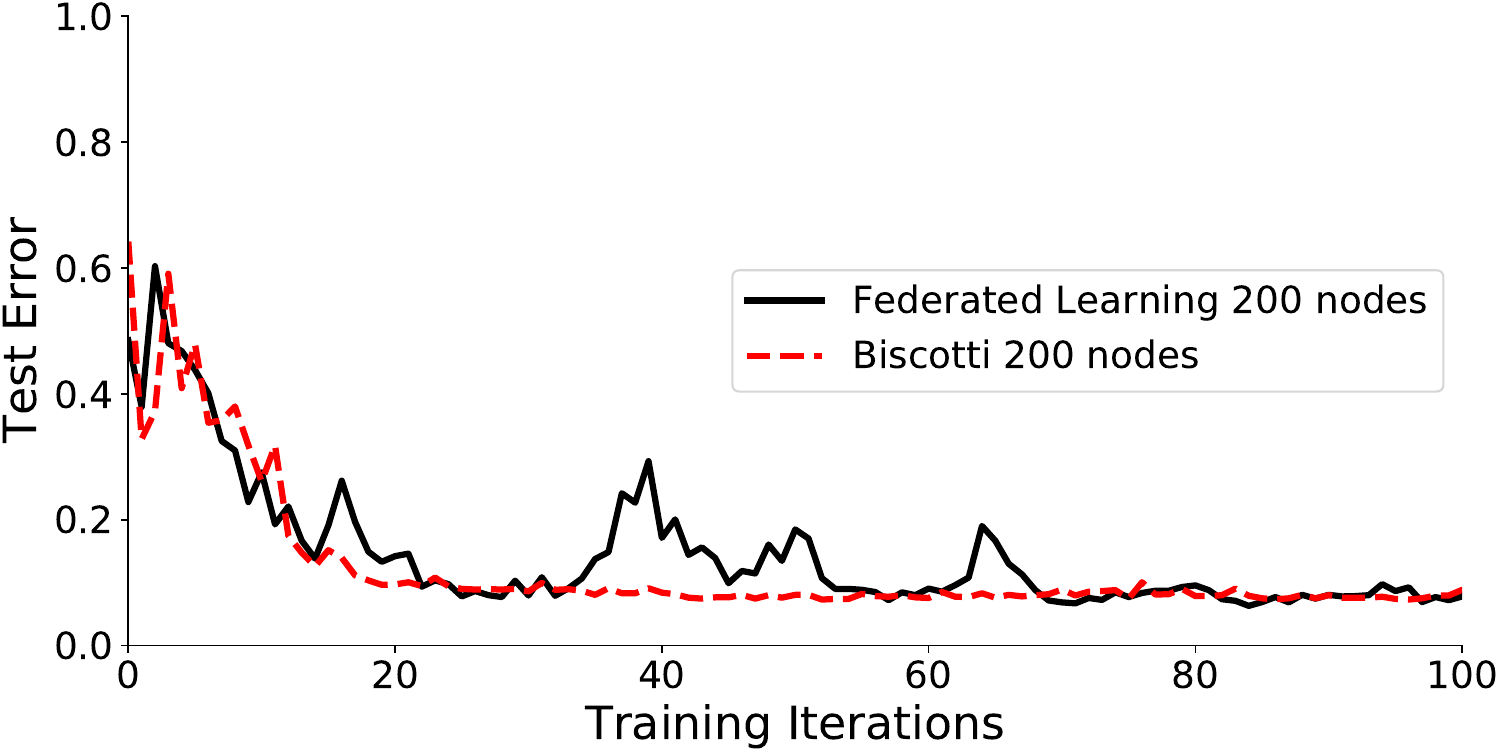}
  \caption{Comparing the iterations to convergence of Biscotti to a
  federated learning baseline.
  }
  \label{fig:evalconv}
\end{figure}

\noindent \textbf{Baseline performance}


%
%


We execute Biscotti in a baseline deployment and compare it to
the original federated learning baseline~\cite{McMahan:2017}. We run Biscotti with 5 noisers, 26 verifiers and 26 aggregators. As demonstrated earlier, these committee sizes provide a guarantee of protecting against an adversary holding 30\% stake from unmasking updates in the noising (Section ~\ref{eval:privacy}) and aggregation stages (Section ~\ref{eval:vrfsecurity}). They also ensure convergence under poisoning from an attack by 30\% of the nodes. (Section ~\ref{eval:poisoning})     
The MNIST dataset~\cite{Lecun:1998} into 200 equal
partitions, each of which was shared with an honest peer on an Azure
cluster of 20 VMs, with each VM hosting 10 peers. These
Biscotti/Federated Learning peers collaborated on training an MNIST
softmax classifier, and after 100 iterations both models approached
the global optimum. To ensure a fair comparison, the number of updates
included in the model each round are kept the same. Federated Learning
selects 35\% nodes at random out of 200 for every round and receives
updates from them while Biscotti includes the first 35\% verified
updates in the block. The convergence rate over time for both systems
is plotted in Figure~\ref{fig:evalconvtime} and the same convergence
over the number of iterations is shown in
Figure~\ref{fig:evalconv}. In this deployment, Biscotti takes about
13.8 times longer than Federated Learning (20.8 minutes vs 266.7 
minutes), yet achieves similar model performance (92\% accuracy)
after 100 iterations.


\begin{figure}[t]
  \centering
  \includegraphics[width=\linewidth]{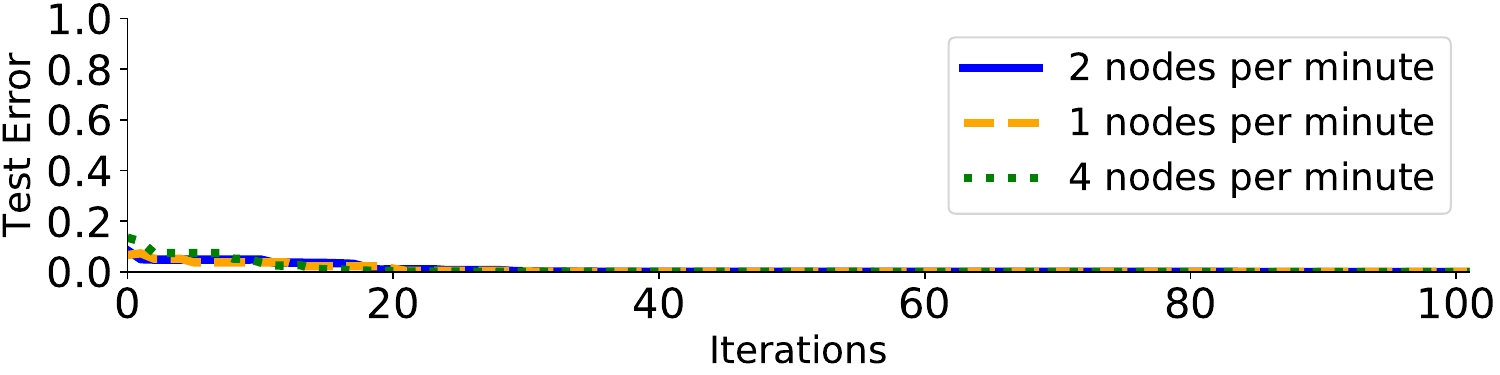}
  \caption{The impact of churn on model performance.
    %
  }
  \label{fig:evalchurn}
\end{figure}

\noindent \textbf{Training with node churn.}
A key feature of Biscotti's P2P design is that it is resilient to node
churn (node joins and failures). For example, the failure of any
Biscotti node does not block or prevent the system from converging.
We evaluate Biscotti's resilience to peer churn by performing a Credit Card
deployment with 100 peers, failing a peer at random at a specified
rate. For each specified time interval, a peer is chosen randomly and
is killed. In the next time interval, a new peer joins the network and
is bootstrapped, maintaining a constant total number of 100
peers. Figure~\ref{fig:evalchurn} shows the rate of convergence for
varying failure rates of 4, 2, and 1 peer(s) failing and joining per
minute. When a verifier or an aggregator fails, Biscotti defaults to the
next iteration after a timeout, so this does not harm convergence.

Even with churn Biscotti is able to make progress towards the global
objective. 
We found that Biscotti is resilient to
churn rates up to 4 nodes per minute (1 node joining and 1 node
failing every 15 seconds).

%% file: discussion.tex
\section{Limitations}
\label{sec:discuss}

\noindent \textbf{KRUM limitations.}  
For KRUM to be
effective against a large number of poisoners, it needs to observe a large
number samples in each round. This may not always be
possible in a decentralized system where there is node churn.
In particular, this may compromise the system because
adversaries may race to contribute a majority of the 
samples. In addition to this, Multi-KRUM could also reject updates of peers that have different data from the rest of the peers e.g only one peer has 1's in the system. We did not face this issue in our experiments because we partitioned the data uniformly. However, Biscotti is compatible with other poisoning detection
approaches and can integrate any poisoning detection mechanism which
uses SGD updates and does not require individual update histories.
For example, in a previous version of Biscotti we used RONI~\cite{Barreno:2010}
to validate updates.


\noindent \textbf{Deep Learning.} We showed that Biscotti is able to
train a softmax and logistic regression model with 7,850 and 25
parameters respectively. But, Biscotti might not scale to a large deep
learning model with millions of parameters due to the communication
overhead. Strategies to reduce this might require learning updates in
a restricted parameter space or compressing model
updates~\cite{Konecny:2016}. We leave the training of deep learning models with
Biscotti to future work.

\noindent \textbf{Leakage from the aggregate model.}
Since there is no differential privacy added to the updates present in
the ledger, Biscotti is vulnerable to attacks that exploit privacy
leakage from the model itself~\cite{Fredrikson:2015,Salem:2018}. Apart
from differential privacy, these attacks can also be mitigated by
adding proper regularization~\cite{Salem:2018,Melis:2018} like dropout,
or they may be irrelevant if a class does not represent an individual's
training data.


\noindent \textbf{Stake limitations.} 
The stake that a client possesses plays a significant role in
determining the chance that a node is selected as a
noiser/verifier/aggregator. Our assumption is that a large stake-holder
will not subvert the system because (1) they accrue more stake by
participating and (2) their stake is tied to a monetary reward at the
end of training. However, a malicious client could pose honest and accrue stake for a pre-defined period of time and then turn malicious to subvert our system. We leave the design of a robust stake mechanism that protects against such attacks to future work.  

%% file: related.tex
\section{Related work}
\label{sec:related}



Biscotti's novelty lies in its ability to simultaneously handle
poisoning attacks and preserve privacy in a P2P multi-party training
context. \vspace{.1em}

\noindent \textbf{Securing ML.}
Similar to KRUM, AUROR~\cite{Shen:2016} and ANTIDOTE~\cite{Rubinstein:2009}
are alternative techniques to defend against poisoning that rely on anomaly
detection. AUROR has been mainly proposed for the model averaging use
case and uses k-means clustering on a subset of important features to
detect anomalies. ANTIDOTE uses a robust PCA detector to protect
against attackers trying to evade anomaly-based methods.

Other defenses like TRIM~\cite{Jagielski:2018} and
RONI~\cite{Barreno:2010} filter out poisoned data from a centralized
training set based on their impact on performance on the dataset. TRIM
trains a model to fit a subset of samples selected at random, and
identifies a training sample as an outlier if the error when fitting
the model to the sample is higher than a threshold. RONI trains a
model with and without a data point and rejects it if it degrades the
performance of the classifier by a certain amount.

Finally, Baracaldo et al.~\cite{Baracaldo:2017} employ data provenance
as a measure against poisoning attacks by tracking the history of the
training data and removing information from anomalous sources.

\noindent \textbf{Poisoning attacks.} 
Apart from label flipping attacks~\cite{Huang:2011} we evaluated,
gradient ascent techniques~\cite{Jagielski:2018,Munoz-Gonzalez:2017}
are a popular way of generating poisoned samples one sample at a time
by solving a bi-level optimization problem. 
Backdoor attacks have also been proposed to make the
classifier misclassify an image if it contains certain pixels or a
backdoor key. Defending against such attacks during training is a hard
and open problem in the literature.

\noindent \textbf{Privacy attacks.}
Shokri et al.~\cite{Shokri:2017} demonstrated a membership inference
attack using shadow model training that learns if a victim's data was
used to train the model. Follow up work~\cite{Salem:2018} showed that
it is quite easy to launch this attack in a black-box setting. In
addition, model inversion attacks~\cite{Fredrikson:2015} have been
proposed to invert class images from the final trained model. However,
we assume that these are only effective if a class represents a
significant chunk of a person's data e.g., facial recognition systems.


Additional security work in federated learning has required that
public key infrastructure exists which
validates the identity of users~\cite{Bonawitz:2017}, but prior sybil
work has shown that relying on public key infrastructure for user
validation is insufficient~\cite{Viswanath:2015, Wang:2016}. 

\noindent \textbf{Privacy-preserving ML.}
Cheng et al. recently proposed the use of TEEs and privacy-preserving
smart contracts to provide security and privacy to multi-party ML
tasks~\cite{Cheng:2018}. But, some TEEs have been found to be
vulnerable~\cite{Foreshadow:2018}. Another solution uses two party
computation~\cite{Mohassel:2017} and encrypts both the model parameters
and the training data when performing multi-party ML.

Differential privacy is often applied to multi-party ML systems to
provide privacy~\cite{Geyer:2017, Abadi:2016, Shokri:2015}. However,
its use introduces privacy-utility tradeoffs. Biscotti uses
differential privacy during update validation, but does not push noise
into the trained model, thus favouring utility.




%% file: conc.tex
\section{Conclusion}
\label{sec:conc}

The emergence of large scale multi-party ML workloads and distributed
ledgers for scalable consensus have produced two rapid and powerful
trends. Biscotti's design lies at their confluence. 
To the best of our knowledge, this is the first system to provide
privacy-preserving peer-to-peer ML through a distributed ledger, while
simultaneously considering poisoning attacks.
And, unlike prior work, 
Biscotti does not rely on trusted execution
environments or specialized hardware.
In our evaluation we demonstrated that Biscotti can coordinate a
collaborative learning process across 100 peers and produces a final
model that is similar in utility to state of the art federated
learning alternatives. We also illustrated its ability to withstand
poisoning and information leakage attacks, and frequent failures and
joining of nodes (one node joining, one node leaving every 15
seconds).

Our Biscotti prototype is open source, runs on commodity hardware, and
interfaces with PyTorch, a popular framework for machine learning.




%% file: appendix.tex
\section{Federated learning and Distributed SGD}
\label{sec:background}

Given a set of training data, a model structure, and a proposed
learning task, ML algorithms \emph{train} an optimal set of
parameters, resulting in a model that optimally performs this task. 
In Biscotti, we assume stochastic gradient descent 
(SGD)~\cite{Bottou:2010} as the optimization algorithm.

In federated learning~\cite{McMahan:2017}, a shared model is updated
through SGD. Each client uses their local training data and their
latest snapshot of the shared model state to compute the optimal update
on the model parameters. The model is then updated and clients update
their local snapshot of the shared model before performing a local
iteration again. The model parameters $w$ are updated at each
iteration $i$ as follows:
\begin{align*}
  w_{t+1} = w_t - \eta_t(\lambda w_t + \frac{1}{b} \sum_{(x_i, y_i)\in
  B_t} \nabla l(w_t, x_i, y_i)) &&&& (1a)
\end{align*}
where $\eta_t$ represents a degrading learning rate, $\lambda$ is a
regularization parameter that prevents over-fitting, $B_t$ represents
a gradient batch of local training data examples $(x_i, y_i)$ of size
$b$ and $\nabla l$ represents the gradient of the loss function. 

SGD is a general learning algorithm that can be used to train a variety
of models, including neural networks~\cite{Bottou:2010}. A typical
heuristic involves running SGD for a fixed number of iterations or
halting when the magnitude of the gradient falls below a threshold.
When this occurs, model training is considered complete and the shared
model state $w_t$ is returned as the optimal model $w^*$.

In a multi-party ML setting federated learning assumes that clients
possess training data that is not identically and independently
distributed (non-IID) across clients. In other words, each client
possesses a subset of the global dataset that contains specific
properties distinct from the global distribution.

When performing SGD across clients with partitioned data sources, we
redefine the SGD update $\Delta_{i,t} w_g$ at iteration $t$ of each
client $i$ to be:
\begin{align*}
  \Delta_{i,t} w_g = \lambda \eta_t w_g + \frac{\eta_t}{b} \sum_{(x,
  y) \in B_{i,t}} \nabla l(w_g, x, y) &&&& (1b)
\end{align*}
where the distinction is that the gradient is computed on a global
model $w_g$, and the gradient steps are taken using a local
batch $B_{i,t}$ of data from client $i$. When all SGD updates are
collected, they are averaged and applied to the model, resulting in a
new global model. The process then proceeds iteratively until
convergence.

To increase privacy guarantees in federated learning, secure
aggregation protocols have been added to the central 
server~\cite{Bonawitz:2017} such that no individual client's SGD update is
directly observable by server or other clients. However, this relies on
a centralized service to perform such an aggregation and does not
provide security against adversarial attacks on ML.

\section{Differentially private stochastic gradient descent}
\label{sec:diffpriv}

\begin{algorithm}[h]
\DontPrintSemicolon
  \KwData{Batch size $b$, Learning rate $\eta_t$, Privacy parameters
  $(\varepsilon , \delta)$, Expected update dimension $d$}
    \KwResult{Precommited noise for an SGD update $\zeta_t$
    $\forall T$}
    \For{iteration $t \in [1 .. T]$}{
      \tcp{Sample noise of length $d$ for each expected sample in batch}
      \For{Example $i \in [1 .. b]$}{
        Sample noise $\zeta_i = \mathcal{N}(0, \sigma^2 I)$
        where $\sigma = \sqrt{2log\frac{1.25}{\delta}}/\epsilon$ 
      }
      $\zeta_t = \frac{\eta_t}{b} \sum_i \zeta_i$ \\
      Commit $\zeta_t$ to the genesis block at column $t$.
  }
  \caption{Precommiting differentially Private noise for SGD, taken
  from Abadi et al.~\cite{Abadi:2016}.
  \label{alg:diffpriv}
  }
\end{algorithm}

We use the concept of $(\epsilon, \delta)$ differential privacy as explained in Abadi et al.~\cite{Abadi:2016}. Each SGD step becomes $(\epsilon, \delta)$ differentially private if we sample normally distributed noise as shown in Algorithm~\ref{alg:diffpriv}. Each
client commits noise to the genesis block for all expected iterations
$T$. We also requires that the norm of the gradients be
clipped to have a maximum norm of 1 so that the noise does not completely obfuscate the gradient.


This precommited noise is designed such that a neutral third party
aggregates a client update $\Delta_{i,t} w_g$ from Equation (1b) and
precommitted noise $\zeta_t$ from Algorithm~\ref{alg:diffpriv} without
any additional information. The noise is generated without any prior
knowledge of the SGD update it will be applied to while retaining the
computation and guarantees provided by prior work. The noisy SGD
update $\widetilde{\Delta_{i,t}} w_g$ follows from aggregation:
\begin{align*}
\widetilde{\Delta_{i,t}} w_g = \Delta_{i,t} w_g + \zeta_t
\end{align*}

\section{Polynomial commitments and verifiable secret sharing}
\label{sec:crypto}

Polynomial Commitments \cite{Aniket:2010} is a scheme that allows commitments to a 
secret polynomial for verifiable secret sharing \cite{Shamir:1979}.
This allows the committer to distribute secret shares for a secret polynomial among a set
of nodes along with 
witnesses that prove in zero-knowledge that each secret share belongs
to the committed polynomial. The polynomial commitment is constructed
as follows:

Given two groups $G_{1}$ and $G_{2}$ with generators $g_{1}$ and $g_{2}$ of prime order $p$ such that there exists an
asymmetric bilinear pairing $e : G_{1} \times  G_{2} \rightarrow G_{T}$ for which the t-SDH
assumption holds, a commitment public key (PK) is generated such that $ PK = \{g, g^{\alpha}, g^{(\alpha)^{2}}, ..., g^{(\alpha)^{t}}\} \in G_{1}^{t+1}  $ where $\alpha$ is the secret key. The committer can create a commitment to a
polynomial $\phi(x) = \sum_{j=0}^{t} \phi_{j}x^{j} $ of degree $t$ using the commitment PK such that:
\begin{align*}
      COMM(PK, \phi(x)) = \prod_{j=0}^{deg(\phi)} (g^{\alpha^{j}})^{\phi_{j}}
\end{align*}

Given a polynomial $\phi(x)$ and a commitment $COMM(\phi(x))$, it is
trivial to
verify whether the commitment was generated using the given polynomial
or not. Moreover, we can multiply two commitments to obtain a commitment to the sum of the polynomials in the commitments by leveraging their homomorphic property: 

\begin{align*}
      COMM(\phi_{1}(x) + \phi_{2}(x) ) = COMM(\phi_{1}(x)) * COMM(\phi_{2}(x))
\end{align*}

Once the committer has has generated $COMM(\phi(x))$, it can carry out a $(n,t)$
- secret sharing scheme to share the polynomial among a set of $n$
participants in such a way that in the recovery phase a subset of
at least $t$
participants can compute the secret polynomial. All secret shares  
$(i, \phi(i))$ shared with the participants are evaluations of the
polynomial at a unique point $i$
and are accompanied by a commitment to a witness polynomial $COMM(\psi_{i}(x))$ 
 such that $\psi_{i}(x) = \frac{\phi(x) - \phi(i)}{x - i}$. By
 leveraging the
divisibility property of the two polynomials $\{\phi(x),\psi(x)\}$ and the bilinear pairing
function $e$, it is trivial to verify that the secret share comes from the
committed polynomial \cite{Aniket:2010}. This is carried out by evaluating
whether the following equality holds:

      \begin{align*}
          e(COMM(\phi_{i}(x)), g_{2}) \stackrel{?}
          {=} e(COMM(\psi_{i}(x)), \frac{g_{2}^{\alpha}}{g_{2}^{i}})e(g_{1}, g_
          {2})^{\Delta w_{i}(i)}
      \end{align*}

      If the above holds, then the share is accepted. Otherwise,
      the share is rejected.

This commitment scheme is unconditionally binding and computationally
hiding
given the Discrete Logarithm assumption holds \cite{Aniket:2010}. This scheme can be
easily extended to provide unconditional hiding by combining it with
another
scheme called Pedersen commitments \cite{Pedersen:1992} however we do
not implement it.  

\section{Verifiable random functions}
\label{sec:vrf}

A verifiable random function $VRF_{sk}(x)$ is a function that takes in as input a random seed $(x)$ and a secret key $(sk)$. Subsequently, it outputs two values: a hash and a proof. The hash output is a hashlen-bit-long value that is uniquely determined by the $sk$ therefore is unique to a peer. The proof allows anyone with the peer's public key (pk) to check that the hash has indeed been generated by a client who holds the private key. Therefore, it provides each client in the system to deterministically produce a  hash that cannot be faked and is unique to the client for that seed value.

In Biscotti, a VRF is used to select a unique committees that provides noise to a peer for protecting its update. A peer uses as inputs to the VRF its own secret key and the SHA-256 hash of the previous block. To select a committee, the hash output of the VRF is input to a consistent hashing procedure.  The hash output from the VRF is mapped to a hash ring where each peer is assigned a space proportional to their stake. (See Figure~\ref{fig:stakering}). The peer in whose portion of the ring the hash lies is selected as part of the committee. The process is repeated until the peer gets a committee of the right size.

\section{Information leakage attacks}
\label{sec:infoleakage}

When doing collaborative learning, each client computes their gradients by back-propagating the loss through the entire network from the last layer to the first layer. The gradient for a layer is computed by using the layer's features and the error from the preceding layer. If the layers are sequential and fully connected, then the output for layer $h_{l+1}$ is computed as follows:

  \begin{align*}
      h_{l+1} = W_{l}*h{l}
  \end{align*}

  where $W_{l}$ is the weight matrix.

  Hence, the gradient of the error with respect to $W_{l}$ is computed as follows:

  \begin{align*}
      \frac{dE}{dW_{l}} = \frac{dE}{dh_{l+1}} * h_{l} 
  \end{align*}

  Note that the gradient of the weights $\frac{dE}{dW_{l}}$ is computed by using the inner products from the layer above and also the features of that particular layer. Therefore, the values of the gradients of the first layer will be proportional to the input features. By exploiting this property, observation of gradient updates can be used to infer feature values, which are in turn based on the participants private training data. 

  In the information leakage attack that we launch on Biscotti, we select the values of the gradient in the first layer that correspond to a certain class, rescale the values to lie between (0,255) and visualize the resulting image.

\section{Proof of stake}
\label{sec:stake}

Blockchain based systems need a mechanism to prevent any arbitrary peer from proposing blocks and extending the chain at the same time. Otherwise, anyone can extend the blockchain and nothing would stop the blockchain from developing forks at a rate equal to the number of users and there will be no consensus eventually. A rate limiting solution is needed to prevent the ledger from being extended infinitely by all the peers. Proof of Work (POW) and Proof of Stake (POS) are two popular solutions to this problem.

Proof of Work uses a puzzle to solve the rate limiting problem but it has its limitations. Peers who want to propose the next block have to solve a hard cryptographic puzzle that is trivial to verify by other peers. The peer that solves the puzzle first gets to propose the next block. The puzzle acts as a rate limiter because by creating new identities (Sybils) peers do not gain any advantage in being the next proposer. In addition to preventing Sybils, it acts as a probabilistic back off mechanism that tries to limit forks in the system. It does not completely eliminate forks and users have to wait for a certain amount of time (6 blocks in Bitcoin) before their transaction is confirmed as accepted. This leads to long wait times (60 minutes for Bitcoin) for transactions to be accepted. Furthermore, it is also energy consuming since it takes a lot of computational power to solve the puzzles. Despite its limitations of long wait times and high energy consumption, POW is a widely used solution. 

To mitigate the problems of Proof of Work, Proof of Stake has been recently proposed as an alternate. POS based systems assign the responsibility of proposing blocks each round to specific peers. The probability that a peer will be selected to propose is proportional to the value (stake) that they have in the system. In cryptocurrencies, the stake of a peer is equal to the amount of money that they have in the system. Based on the distribution of stake at that particular time, a predefined algorithm is used to choose a peer/subset of peers that is responsible for proposing the next block.  The algorithm ensures that that at any time a group of malicious nodes holding a certain amount of stake in the system cannot act maliciously and take over the system. All other users observe the protocol messages, which allows them to learn the agreed-upon block.    

For cryptocurrencies, Algorand~\cite{Gilad:2017} is a popular proposal for a proof of stake based system. In Algorand, each peer is assigned a priority for two particular roles: block proposal and block selection. To determine their priority for a particular role, each peer runs a verifiable random function (VRF's) (see ~\ref{sec:vrf}) using their private key, the role (selection/proposal) and a random seed which is public information on the blockchain. The VRF outputs a pseudo-random hash value that is passed by the user through cryptographic sortition to determine their influence in proposing or selecting a block for that particular round. At a high level, the cryptographic sortition is a random algorithm that assigns each user a priority such that the priority assigned is proportional to the user's account balance. If the user's priority is above a threshold, they are selected for that particular role.
Subsequently, all users assigned a proposal role, propose a block  based on the user's priority. To reach consensus on a single block proposal, peers assigned the selection role agree on one proposal for the next block using a multi-step Byzantine Agreement protocol. In each step, the peers responsible for that step vote for a proposal until in some step enough users have agreed on a proposal. This proposal becomes the next block in the chain.    

Similar to Algorand, Biscotti uses verifiable random functions to assign roles to peers in the system. However, instead of using the cryptographic sortition algorithm, Biscotti uses a consistent hashing protocol described in Section ~\ref{sec:peerselection} to select a peer for a role such that the probability of getting selected for a role is proportional to the peer's stake. In Biscotti, we define stake to be the reputation that a peer acquires over the training process by contributing updates.  

%% file: paper.bbl
\begin{thebibliography}{10}

\bibitem{Coniks}
{A CONIKS Implementation in Golang}.
\newblock \url{https://github.com/coniks-sys/coniks-go}, 2018.

\bibitem{Kyber}
{DEDIS Advanced Crypto Library for Go}.
\newblock \url{https://github.com/dedis/kyber}, 2018.

\bibitem{gopython}
{go-python}.
\newblock \url{https://github.com/sbinet/go-python}, 2018.

\bibitem{Abadi:2016}
{\sc Abadi, M., Chu, A., Goodfellow, I., McMahan, B., Mironov, I., Talwar, K.,
  and Zhang, L.}
\newblock Deep learning with differential privacy.
\newblock In {\em 23rd ACM Conference on Computer and Communications
  Security\/} (2016), CCS.

\bibitem{Azaria:2016}
{\sc Azaria, A., Ekblaw, A., Vieira, T., and Lippman, A.}
\newblock Medrec: Using blockchain for medical data access and permission
  management.
\newblock OBD '16.

\bibitem{Bagdasaryan:2018}
{\sc {Bagdasaryan}, E., {Veit}, A., {Hua}, Y., {Estrin}, D., and {Shmatikov},
  V.}
\newblock {How To Backdoor Federated Learning}.
\newblock {\em ArXiv e-prints\/} (2018).

\bibitem{Baracaldo:2017}
{\sc Baracaldo, N., Chen, B., Ludwig, H., and Safavi, J.~A.}
\newblock Mitigating poisoning attacks on machine learning models: A data
  provenance based approach.
\newblock In {\em Proceedings of the 10th ACM Workshop on Artificial
  Intelligence and Security\/} (2017), AISec.

\bibitem{Barreno:2010}
{\sc Barreno, M., Nelson, B., Joseph, A.~D., and Tygar, J.~D.}
\newblock {The Security of Machine Learning}.
\newblock {\em Machine Learning 81}, 2 (2010).

\bibitem{Bhagoji:2019}
{\sc Bhagoji, A.~N., Chakraborty, S., Mittal, P., and Calo, S.}
\newblock Analyzing federated learning through an adversarial lens.
\newblock In {\em Proceedings of the 36th International Conference on Machine
  Learning\/} (2019).

\bibitem{Biggio:2012}
{\sc Biggio, B., Nelson, B., and Laskov, P.}
\newblock {Poisoning Attacks Against Support Vector Machines}.
\newblock In {\em Proceedings of the 29th International Coference on
  International Conference on Machine Learning\/} (2012), ICML.

\bibitem{Blanchard:2017}
{\sc Blanchard, P., El~Mhamdi, E.~M., Guerraoui, R., and Stainer, J.}
\newblock Machine learning with adversaries: Byzantine tolerant gradient
  descent.
\newblock In {\em Advances in Neural Information Processing Systems 30}, NIPS.
  2017.

\bibitem{Bonawitz:2017}
{\sc Bonawitz, K., Ivanov, V., Kreuter, B., Marcedone, A., McMahan, H.~B.,
  Patel, S., Ramage, D., Segal, A., and Seth, K.}
\newblock Practical secure aggregation for privacy-preserving machine learning.
\newblock In {\em Proceedings of the 2017 ACM SIGSAC Conference on Computer and
  Communications Security\/} (2017), CCS.

\bibitem{Bottou:1999}
{\sc Bottou, L.}
\newblock On-line learning in neural networks.
\newblock Cambridge University Press, 1999.

\bibitem{Bottou:2010}
{\sc Bottou, L.}
\newblock {\em {Large-Scale Machine Learning with Stochastic Gradient
  Descent}}.
\newblock COMPSTAT '10. 2010.

\bibitem{Castro:1999}
{\sc Castro, M., and Liskov, B.}
\newblock Practical byzantine fault tolerance.
\newblock In {\em Proceedings of the Third Symposium on Operating Systems
  Design and Implementation\/} (1999), OSDI '99.

\bibitem{Cheng:2018}
{\sc {Cheng}, R., {Zhang}, F., {Kos}, J., {He}, W., {Hynes}, N., {Johnson}, N.,
  {Juels}, A., {Miller}, A., and {Song}, D.}
\newblock Ekiden: A platform for confidentiality-preserving, trustworthy, and
  performant smart contract execution.
\newblock {\em ArXiv e-prints\/} (2018).

\bibitem{Dean:2012}
{\sc Dean, J., Corrado, G.~S., Monga, R., Chen, K., Devin, M., Le, Q.~V., Mao,
  M.~Z., Ranzato, M., Senior, A., Tucker, P., Yang, K., and Ng, A.~Y.}
\newblock Large scale distributed deep networks.
\newblock In {\em Proceedings of the 25th International Conference on Neural
  Information Processing Systems - Volume 1\/} (2012), NIPS.

\bibitem{Dua:2017}
{\sc Dheeru, D., and Karra~Taniskidou, E.}
\newblock {UCI} machine learning repository, 2017.

\bibitem{Douceur:2002}
{\sc Douceur, J.~J.}
\newblock The sybil attack.
\newblock IPTPS '02.

\bibitem{Dwork:2014}
{\sc Dwork, C., and Roth, A.}
\newblock {The Algorithmic Foundations of Differential Privacy}.
\newblock {\em Foundations and Trends in Theoretical Computer Science 9}, 3-4
  (2014).

\bibitem{Fredrikson:2015}
{\sc Fredrikson, M., Jha, S., and Ristenpart, T.}
\newblock {Model Inversion Attacks That Exploit Confidence Information and
  Basic Countermeasures}.
\newblock In {\em Proceedings of the 2015 ACM SIGSAC Conference on Computer and
  Communications Security\/} (2015), CCS.

\bibitem{Geyer:2017}
{\sc Geyer, R.~C., Klein, T., and Nabi, M.}
\newblock Differentially private federated learning: {A} client level
  perspective.
\newblock {\em NIPS Workshop: Machine Learning on the Phone and other Consumer
  Devices\/} (2017).

\bibitem{Gilad:2017}
{\sc Gilad, Y., Hemo, R., Micali, S., Vlachos, G., and Zeldovich, N.}
\newblock Algorand: Scaling byzantine agreements for cryptocurrencies.
\newblock In {\em Proceedings of the 26th Symposium on Operating Systems
  Principles\/} (2017), SOSP.

\bibitem{Hitaj:2017}
{\sc Hitaj, B., Ateniese, G., and P{\'{e}}rez{-}Cruz, F.}
\newblock {Deep Models Under the {GAN:} Information Leakage from Collaborative
  Deep Learning}.
\newblock In {\em Proceedings of the 2017 ACM SIGSAC Conference on Computer and
  Communications Security\/} (2017), CCS.

\bibitem{Huang:2011}
{\sc Huang, L., Joseph, A.~D., Nelson, B., Rubinstein, B.~I., and Tygar, J.~D.}
\newblock {Adversarial Machine Learning}.
\newblock In {\em Proceedings of the 4th ACM Workshop on Security and
  Artificial Intelligence\/} (2011), AISec.

\bibitem{Jagielski:2018}
{\sc Jagielski, M., Oprea, A., Biggio, B., Liu, C., Nita{-}Rotaru, C., and Li,
  B.}
\newblock Manipulating machine learning: Poisoning attacks and countermeasures
  for regression learning.
\newblock In {\em 2018 {IEEE} Symposium on Security and Privacy, {SP} 2018,
  Proceedings, 21-23 May 2018, San Francisco, California, {USA}}.

\bibitem{Aniket:2010}
{\sc Kate, A., and Zaverucha, Gregory M.and~Goldberg, I.}
\newblock Constant-size commitments to polynomials and their applications.

\bibitem{Aggelos:2017}
{\sc Kiayias, A., Russell, A., David, B., and Oliynykov, R.}
\newblock Ouroboros: A provably secure proof-of-stake blockchain protocol.
\newblock In {\em Advances in Cryptology -- CRYPTO 2017\/} (2017).

\bibitem{Konecny:2016}
{\sc Konečný, J., McMahan, H.~B., Yu, F.~X., Richtarik, P., Suresh, A.~T.,
  and Bacon, D.}
\newblock Federated learning: Strategies for improving communication
  efficiency.
\newblock In {\em NIPS Workshop on Private Multi-Party Machine Learning\/}
  (2016).

\bibitem{Lecun:1998}
{\sc Lecun, Y., Bottou, L., Bengio, Y., and Haffner, P.}
\newblock Gradient-based learning applied to document recognition.
\newblock {\em Proceedings of the IEEE 86}, 11 (1998).

\bibitem{Maxwell:2018}
{\sc Maxwell, G., Poelstra, A., Seurin, Y., and Wuille, P.}
\newblock Simple schnorr multi-signatures with applications to bitcoin.
\newblock Cryptology ePrint Archive, Report 2018/068, 2018.

\bibitem{McMahan:2017}
{\sc McMahan, H.~B., Moore, E., Ramage, D., Hampson, S., and y~Arcas, B.~A.}
\newblock {Communication-Efficient Learning of Deep Networks from Decentralized
  Data}.
\newblock In {\em Proceedings of the 20th International Conference on
  Artificial Intelligence and Statistics\/} (2017), AISTATS.

\bibitem{Melis:2018}
{\sc {Melis}, L., {Song}, C., {De Cristofaro}, E., and {Shmatikov}, V.}
\newblock {Inference Attacks Against Collaborative Learning}.
\newblock {\em ArXiv e-prints\/} (2018).

\bibitem{Micali:1999}
{\sc Micali, S., Vadhan, S., and Rabin, M.}
\newblock Verifiable random functions.
\newblock In {\em Proceedings of the 40th Annual Symposium on Foundations of
  Computer Science\/} (1999), FOCS.

\bibitem{Miller:2014}
{\sc Miller, A., Juels, A., Shi, E., Parno, B., and Katz, J.}
\newblock Permacoin: Repurposing bitcoin work for data preservation.
\newblock In {\em Proceedings of the 2014 IEEE Symposium on Security and
  Privacy\/} (2014), S\&P.

\bibitem{Mohassel:2017}
{\sc Mohassel, P., and Zhang, Y.}
\newblock Secureml: A system for scalable privacy-preserving machine learning,
  2017.

\bibitem{Munoz-Gonzalez:2017}
{\sc Mu\~{n}oz Gonz\'{a}lez, L., Biggio, B., Demontis, A., Paudice, A.,
  Wongrassamee, V., Lupu, E.~C., and Roli, F.}
\newblock Towards poisoning of deep learning algorithms with back-gradient
  optimization.
\newblock In {\em Proceedings of the 10th ACM Workshop on Artificial
  Intelligence and Security 2017}.

\bibitem{Nakamoto:2009}
{\sc Nakamoto, S.}
\newblock {Bitcoin: A peer-to-peer electronic cash system}.

\bibitem{Narula:2018}
{\sc Narula, N., Vasquez, W., and Virza, M.}
\newblock zkledger: Privacy-preserving auditing for distributed ledgers.
\newblock In {\em 15th {USENIX} Symposium on Networked Systems Design and
  Implementation\/} (2018), NSDI.

\bibitem{Nelson:2008}
{\sc Nelson, B., Barreno, M., Chi, F.~J., Joseph, A.~D., Rubinstein, B. I.~P.,
  Saini, U., Sutton, C., Tygar, J.~D., and Xia, K.}
\newblock {Exploiting Machine Learning to Subvert Your Spam Filter}.
\newblock In {\em Proceedings of the 1st Usenix Workshop on Large-Scale
  Exploits and Emergent Threats\/} (2008), LEET.

\bibitem{Pytorch}
{\sc Paszke, A., Gross, S., Chintala, S., Chanan, G., Yang, E., DeVito, Z.,
  Lin, Z., Desmaison, A., Antiga, L., and Lerer, A.}
\newblock Automatic differentiation in pytorch.
\newblock In {\em NIPS Autodiff Workshop\/} (2017).

\bibitem{Pedersen:1992}
{\sc Pedersen, T.~P.}
\newblock Non-interactive and information-theoretic secure verifiable secret
  sharing.
\newblock In {\em Proceedings of the 11th Annual International Cryptology
  Conference\/} (1992), CRYPTO.

\bibitem{Recht:2011}
{\sc Recht, B., Re, C., Wright, S., and Niu, F.}
\newblock Hogwild: A lock-free approach to parallelizing stochastic gradient
  descent.
\newblock In {\em Advances in Neural Information Processing Systems 24}, NIPS.
  2011.

\bibitem{Rubinstein:2009}
{\sc Rubinstein, B.~I., Nelson, B., Huang, L., Joseph, A.~D., Lau, S.-h., Rao,
  S., Taft, N., and Tygar, J.~D.}
\newblock {ANTIDOTE: Understanding and Defending Against Poisoning of Anomaly
  Detectors}.
\newblock In {\em Proceedings of the 9th ACM SIGCOMM Conference on Internet
  Measurement\/} (2009), IMC.

\bibitem{Salem:2018}
{\sc Salem, A., Zhang, Y., Humbert, M., Fritz, M., and Backes, M.}
\newblock Ml-leaks: Model and data independent membership inference attacks and
  defenses on machine learning models.

\bibitem{Schnorr:1990}
{\sc Schnorr, C.~P.}
\newblock Efficient identification and signatures for smart cards.
\newblock In {\em Advances in Cryptology --- CRYPTO' 89 Proceedings\/} (1990).

\bibitem{Shamir:1979}
{\sc Shamir, A.}
\newblock How to share a secret.
\newblock {\em Communications of the ACM\/} (1979).

\bibitem{Shen:2016}
{\sc Shen, S., Tople, S., and Saxena, P.}
\newblock Auror: Defending against poisoning attacks in collaborative deep
  learning systems.
\newblock In {\em Proceedings of the 32nd Annual Conference on Computer
  Security Applications\/} (2016), ACSAC.

\bibitem{Shokri:2015}
{\sc Shokri, R., and Shmatikov, V.}
\newblock Privacy-preserving deep learning.
\newblock In {\em Proceedings of the 2015 ACM SIGSAC Conference on Computer and
  Communications Security\/} (2015), CCS.

\bibitem{Shokri:2017}
{\sc Shokri, R., Stronati, M., Song, C., and Shmatikov, V.}
\newblock Membership inference attacks against machine learning models.
\newblock In {\em {IEEE} Symposium on Security and Privacy\/} (2017), {IEEE}
  Computer Society, pp.~3--18.

\bibitem{Foreshadow:2018}
{\sc Van~Bulck, J., Minkin, M., Weisse, O., Genkin, D., Kasikci, B., Piessens,
  F., Silberstein, M., Wenisch, T.~F., Yarom, Y., and Strackx, R.}
\newblock Foreshadow: Extracting the keys to the {Intel SGX} kingdom with
  transient out-of-order execution.
\newblock In {\em Proceedings of the 27th {USENIX} Security Symposium\/}
  (2018), USENIX SEC.

\bibitem{Viswanath:2015}
{\sc Viswanath, B., Bashir, M.~A., Zafar, M.~B., Bouget, S., Guha, S., Gummadi,
  K.~P., Kate, A., and Mislove, A.}
\newblock Strength in numbers: Robust tamper detection in crowd computations.
\newblock In {\em {Proceedings of the 3rd ACM Conference on Online Social
  Networks}\/} ({2015}), COSN.

\bibitem{Wang:2016}
{\sc Wang, G., Wang, B., Wang, T., Nika, A., Zheng, H., and Zhao, B.~Y.}
\newblock Defending against sybil devices in crowdsourced mapping services.
\newblock In {\em Proceedings of the 14th Annual International Conference on
  Mobile Systems, Applications, and Services\/} (2016), MobiSys.

\end{thebibliography}
